\pdfoutput=1
\documentclass[11pt,a4paper]{article}

\usepackage{times,latexsym}
\usepackage{url}
\usepackage[T1]{fontenc}
\usepackage[font=small]{caption}
\usepackage{microtype}
\usepackage{graphicx}
\usepackage{subfigure}
\usepackage{booktabs} 
\usepackage{longtable}
\usepackage{adjustbox}
\usepackage{float}
\usepackage{graphicx}

\usepackage[dvipsnames]{xcolor}

\usepackage{hyperref}
\usepackage{xspace}

\usepackage[acceptedWithA]{tacl2021}

\usepackage{amsmath}
\usepackage{amssymb}
\usepackage{mathtools}
\usepackage{amsthm}
\usepackage{multirow}
\usepackage{bbm}
\usepackage{enumitem}
\usepackage[capitalize,noabbrev]{cleveref}
\crefname{section}{\S}{\S\S}
\crefname{table}{Tab.}{Tab.}
\crefname{figure}{Fig.}{}
\crefname{algorithm}{Alg.}{}
\crefname{equation}{Eq.}{Eq.}
\crefname{appendix}{App.}{}
\crefname{theorem}{Theorem}{}
\crefname{prop}{Proposition}{}
\crefname{cor}{Corollary}{}
\crefname{observation}{Observation}{}
\crefname{assumption}{Assumption}{}
\crefname{hypothesis}{Hyp.}{Hypotheses}
\crefformat{section}{\S#2#1#3}
\newcommand{\defeq}[0]{\mathrel{\stackrel{\textnormal{\tiny def}}{=}}}

\theoremstyle{plain}
\newtheorem{theorem}{Theorem}[section]
\newtheorem{proposition}[theorem]{Proposition}
\newtheorem{myexample}[theorem]{Example}
\newtheorem{sampler}[theorem]{Algorithm}

\newtheorem{hypothesis}[theorem]{Hypothesis}

\newtheorem{definition}[theorem]{Definition}

\theoremstyle{remark}

\newcommand{\ent}{\mathrm{H}}

\newcommand{\defn}[1]{\textbf{#1}}

\newcommand{\yy}{\boldsymbol{y}}

\newcommand{\calY}{\mathcal{Y}}

\newcommand{\vtheta}{{\boldsymbol \theta}}

\newcommand{\vocab}{\mathcal{V}}
\newcommand{\vocabeos}{\overline{\mathcal{V}}}
\newcommand{\eos}{\textsc{eos}\xspace}
\newcommand{\bos}{\textsc{bos}\xspace}
\newcommand{\ptrue}{\widetilde{p}}
\newcommand{\pmodel}{\widehat{p}}
\newcommand{\emp}{p}
\newcommand{\model}{\pmodel}

\newcommand{\rep}{\textsc{rep}\xspace}
\newcommand{\response}[1]{\vspace{3pt}\hrule\vspace{3pt}\textbf{#1:}}
\definecolor{darkblue}{rgb}{0.0,0.0,0.5}
\definecolor{purple}{rgb}{0.5,0.0,0.5}
\newcommand{\db}[1]{\textcolor{darkblue}{\bf\scriptstyle \selectfont \,(\pm#1)}}
\newcommand{\ddel}[1]{\textcolor{purple}{\bf\scriptstyle \selectfont \,(#1)}}
\newcommand{\citeposs}[1]{\citeauthor{#1}'s (\citeyear{#1})}

\newcommand{\boldY}{\boldsymbol{Y}}

\usepackage[disable]{todonotes}
\makeatletter
\newcommand*\iftodonotes{\if@todonotes@disabled\expandafter\@secondoftwo\else\expandafter\@firstoftwo\fi}  
\makeatother

\newcommand{\note}[4][]{\todo[author=#2,color=#3,size=\scriptsize,fancyline,caption={},#1]{#4}} 
\newcommand{\ryan}[2][]{\note[#1]{ryan}{violet!40}{#2}}
\newcommand{\clara}[2][]{\note[#1]{clara}{teal!60}{#2}}

\newcommand{\ucambridge}{2}
\newcommand{\ethz}{1}
\usepackage{inconsolata}  

\title{Locally Typical Sampling}

\author{
 Clara Meister$^{\ethz}$~\;~Tiago Pimentel$^{\ucambridge}$~\;~ Gian Wiher$^{\ethz}$~\;~Ryan Cotterell$^{\ethz,\ucambridge}$ \\
   $^{\ethz}$ETH Zürich~\;~ $^{\ucambridge}$University of Cambridge\\
  \texttt{\href{mailto:clara.meister@inf.ethz.ch}{clara.meister@inf.ethz.ch}}~\;~\texttt{\href{mailto:tp472@cam.ac.uk}{tp472@cam.ac.uk}} \\
   \texttt{\href{mailto:gian.wiher@inf.ethz.ch}{gian.wiher@inf.ethz.ch}}~\;~\texttt{\href{mailto:ryan.cotterell@inf.ethz.ch}{ryan.cotterell@inf.ethz.ch}}
}

\date{}

\begin{document}
\maketitle

\begin{abstract}



Today's probabilistic language generators fall short when it comes to producing coherent and fluent text despite the fact that the underlying models perform well under standard metrics, e.g., perplexity.  
This discrepancy has puzzled the language generation community for the last few years. 
In this work, we posit that the abstraction of natural language generation as a discrete stochastic process---which allows for an information-theoretic analysis---can provide new insights into the behavior of probabilistic language generators, e.g., why high-probability texts can be dull or repetitive. 
Humans use language as a means of communicating information, aiming to do so in a simultaneously efficient and error-minimizing manner; in fact, psycholinguistics research suggests humans choose each word in a string with this subconscious goal in mind.
We formally define the set of strings that meet this criterion: those for which each word has an information content close to the \emph{expected} information content, i.e., the conditional entropy of our model. 
We then propose a simple and efficient procedure for enforcing this criterion when generating from probabilistic models, 
which we call \defn{locally typical sampling}.
Automatic and human evaluations show that, in comparison to nucleus and top-$k$ sampling, locally typical sampling offers competitive performance (in both abstractive summarization and story generation) in terms of quality while consistently reducing degenerate repetitions.\looseness=-1

\end{abstract}

\section{Introduction}
Modern probabilistic models have repeatedly demonstrated their prowess at modeling natural language, placing high probability on held-out corpora from many different domains \cite{NEURIPS2020_GPT3,chinchilla,palm}.
Yet when used as text generators, their performance is far from perfect.
One of the largest determinants of the generated text's quality is the choice of \defn{decoding strategy}---i.e., the decision rule used to extract strings from a model.
Perhaps surprisingly, for many language generation tasks, decoding strategies which aim to find the highest-probability strings produce text that is undesirable \cite{holtzman_curious_2020,see-etal-2019-massively,eikema_is_2020,zhang_trading_2020,delucia_decoding_2020}. 
For instance, \citet{stahlberg_nmt_2019} report that in their neural machine translation experiments, the highest-probability string is usually the empty string.
On the other hand, stochastic strategies, which take random samples from the model, often lead to text with better qualitative properties \cite{fan_hierarchical_2018,holtzman_curious_2020,BasuRKV2021}.
However, stochastic strategies still have a host of other problems, while not entirely dispensing with those seen in maximization-based  approaches.\footnote{While maximization-based strategies can produce text which is generic or degenerate, stochastic strategies occasionally produce nonsensical text. Both types of strategies  tend to eventually fall into repetitive loops.\looseness=-1}\looseness=-1


At first glance, it is unintuitive that high-probability strings are often neither desirable nor human-like. 
Due to this pathology, a number of works have concluded that there must be faults in the training objective or architecture of the probabilistic models behind  language generators \citep[][\emph{inter alia}]{welleck2019neural,jian-knowledge-enhanced-2020,li-etal-2020-dont}.
Yet, this conclusion is at odds with these models' performance in terms of other metrics.
The fact that modern models can place high probability on held-out text suggests that they provide good estimates (in at least some aspects) of the probability distribution underlying human language.
We posit that looking at language generation through an information-theoretic lens may shed light on this paradox.\looseness=-1

Communication via natural language can intuitively be cast in information-theoretic terms.
Indeed, there is a long history of studying language through the lens of information theory \citep[][\emph{inter alia}]{shannon1948mathematical,shannon1951prediction,hale2001probabilistic,Piantadosi3526,pimentel-2020-phonotactic}. 
In this paradigm, linguistic strings are messages used to convey information, and their information content can be quantified as a function of their probability of being uttered---often driven by context. 
Assuming that humans use language in order to transmit information in an efficient yet robust manner \cite{zaslavsky2018efficient,gibson2019efficiency}, the subset of strings typically used by humans should encode information at some (perhaps near-optimal) rate.\footnote{Information rate may be defined with respect to time (as is the case with spoken language) or with respect to a specific linguistic unit, such as a word (as is the case with text).}
In fact, prior works studying the uniform information density hypothesis \citep{levy2007speakers,mahowald2013info} empirically observed this property in humans' use of natural language.

These insights lead us to re-think what it means to be a probabilistic language generator. 
First, we contend that language generators, in some cases, can be thought of as discrete stochastic processes.
This in turn, allows us to cleanly define typicality (and the typical set) for these processes.
We argue, however, that due to discrepancies between the model behind these generators and the true distribution over natural language strings, directly sampling from the typical set is not a good idea. 
Indeed, for language generators that do not use an end-of-string ($\eos$) state, this is exactly what is done by ancestral sampling---a decoding strategy not known for providing high-quality text.
Inspired by research on human sentence processing, we then define the more restrictive notion of \emph{local} typicality, and argue that if we want text generated from a model to be ``human-like,'' we should perhaps enforce this information-theoretic criterion in generations ourselves. 
To this end, we develop a new algorithm, which we call \textbf{locally typical sampling}. 
Concretely, we hypothesize that for text to be perceived as natural, each word should have an information content close to its \emph{expected} information content given prior context. 
When sampling from probabilistic language generators, we should limit our options to strings that adhere to this property. 
In experiments on abstractive summarization and story generation, we observe that, compared to nucleus and top-$k$ sampling: (i) locally typical sampling reduces the number of degenerate repetitions, giving a \rep value \cite{welleck2019neural} on par with human text, and (ii) text generated using typical sampling is generally closer in quality to that of human text.\footnote{An implementation of typical sampling can be found in the \href{https://huggingface.co/}{Hugging Face's Transformers library \citep{wolf-etal-2020-transformers}}.\looseness=-1}

\section{Two Views of Language Modeling}\label{sec:two-views}

In this work, we discuss language models\footnote{Here we use the term language model to refer to any (valid) probability distribution over natural language strings. We subsequently specify the necessary conditions for validity. 
Note that this distribution may also be conditioned on an input.} in an information-theoretic light. Our first step towards this goal is to re-frame their presentation.
Concretely, we put forth that there are actually two lenses through which we can view language modeling productively.
Under the traditional lens, we can think of a language model as a distribution over full strings: a language model constitutes the distribution of a single string-valued random variable. 
Under an alternative lens, we can think of a language model as a discrete stochastic process: a collection of indexed random variables.
We compare and contrast these views formally, and
then show how to use the language process view  to derive a new sampling algorithm in \cref{sec:sampling}.\looseness=-1

\subsection{A Single String-Valued Random Variable}\label{sec:language_model}

We codify the traditional view of language modeling in the following definition. Let $\vocab$ be an alphabet---a non-empty, finite set.

\begin{definition}[Language Model]\label{defin:language-model}
A \defn{language model} $p$ is a probability distribution over all strings $\yy \in \vocab^*$.\footnote{The Kleene closure of a set $\vocab$ is defined as $\vocab^* \defeq \bigcup_{n=0}^\infty \vocab^n$.\looseness=-1}
Under this view, we can think of a language model as describing a single $\vocab^*$-valued random variable. 
\end{definition}
Under \Cref{defin:language-model}, it is common to express a language model in the following factorized form\looseness=-1%
\begin{equation}\label{eq:local-norm}
    p(\yy = y_1\cdots y_T) = \prod_{t=1}^T p(y_t \mid \yy_{<t})
\end{equation}
where we define $\yy_{<t} \defeq \langle y_0, \dots, y_{t-1}\rangle$
with the padding $y_0=\bos$ as a distinguished beginning-of-string symbol and $Y_T=\eos$ as a distinguished end-of-string symbol.
Through the chain rule of probability, we can \emph{always} factorize a model as in \cref{eq:local-norm}.
The process which produces such a factorization is called
\defn{local normalization}.\footnote{
The ubiquity of \Cref{eq:local-norm} has led some authors to \emph{defining} language models in the locally normalized form, even though globally normalized language models are also perfectly fine to consider \cite{goyal-etal-2019-empirical}.}
However, with local normalization, we encounter a subtlety: one has to define each conditional probability $p(y_t \mid \yy_{<t})$ not over $\vocab$, but rather over the augmented set $\vocabeos \defeq \vocab \cup \{\eos\}$, i.e., where we have added the distinguished end-of-string symbol $\eos$. 
Why? Because \emph{without} $\eos$, it would be impossible to normalize the language model, i.e., have it sum to $1$.\footnote{Some authors erroneously omit $\eos$ from their definition. However, we \emph{require} a distinguished symbol $\eos$ to be able to locally normalize the language model and make it a valid probability distribution.}

\subsection{A Discrete Stochastic Process} \label{sec:process}
Interestingly, 
the factorization in \cref{eq:local-norm} suggests
that we might view language models, not as a single string-valued random variable, but rather as a collection of random variables $\{Y_t\}_{t=1}^\infty$, i.e., as a discrete stochastic process.\footnote{This process is discrete both in time and in value.}
Under this view, we arrive at the following definition of what we term a language \emph{process}, to distinguish it from the definition of a language model given above.\looseness=-1
\begin{definition}[Language Process]\label{def:lm-stochastic-process}
A \defn{language process} over $\vocab$ is a discrete stochastic process $\boldY = \{Y_t\}_{t=1}^\infty$ where each $Y_t$ is $\vocabeos$-valued.
The process is described by a distribution $p$, and we denote its conditional distribution as $p(Y_t = y_t \mid \boldY_{<t} = \yy_{<t})$  for $t > 0$. In slight abuse of notation but out of convention, we take $Y_{t}$ for $t \leq 0$ to be $\bos$, i.e., conditioning $p$ on just $\bos$ signifies the initial distribution of the process.\looseness=-1
\end{definition}
 \Cref{def:lm-stochastic-process} is very generic.
In words, it just says that a language process is any discrete process where we sample a new word\footnote{One could just as easily define a language process over subwords, morphemes or characters.} given the previously sampled words.
The first question that naturally comes to mind is when the definitions of a language model and a language process coincide.
As it turns out, there is a simple answer.\looseness=-1
\begin{definition}[Tightness]
Let $\boldY = \{Y_t\}_{t=1}^\infty$ be a language process over alphabet $\vocab$ with distribution $p$.
A language process is \defn{tight} \cite{booth1973} if and only if \looseness=-1
\begin{align} \label{eq:tightness}
    \sum_{\yy \in (\vocab^* \otimes \{\eos\})} \prod_{t=1}^{|\yy|} p(&Y_t = y_t \mid \boldY_{<t} = \yy_{<t}) 
    = 1
\end{align}
where $A \otimes B \defeq \{ \boldsymbol{a}\boldsymbol{b} \mid \boldsymbol{a} \in A, \boldsymbol{b} \in B\}$.
\end{definition}
In words, tightness says that a language process must not leak probability mass to infinite strings\clara{Update: I change utterance to string everywhere, since string is the more commonly used term}. 
Because a language model must be a (valid) probability distribution, it must also be tight.\looseness=-1

\begin{proposition}\label{prop:conversion}
Let $\boldY = \{Y_t\}_{t=1}^\infty$ be a language process over alphabet $\vocab$ with distribution $p$ and 
let $p_t \defeq \frac{\sum_{\yy \in (\vocab^{t-1} \otimes \{\eos\})} \prod_{i=1}^{|\yy|} p(Y_i = y_i  \mid\boldY_{<i} = \yy_{<i})}{\sum_{\yy \in \vocab^{t-1}} \prod_{i=1}^{|\yy|} p(Y_i = y_i  \mid\boldY_{<i} = \yy_{<i})}$.
Then $\boldY$ is tight if and only if $p_t = 1$ for some $0 < t < \infty$ or $\sum_{t=1}^\infty p_t \rightarrow \infty$.
\end{proposition}
\begin{proof}
Note that $p_t$ is the probability of sampling $\eos$ at \emph{exactly} step $t$ given that the history of the string is of length $(t-1)$.
\begin{itemize}[leftmargin=*]
\item \textbf{Case 1:}
Suppose $p_t = 1$ for some $0 < t < \infty$.
Then, $\boldY$ is clearly tight as no probability mass
is leaked to strings beyond length $t$, where $t < \infty$.\looseness=-1

\item \textbf{Case 2:}
Now suppose $p_t < 1$ for all $t$.
In this case, we have that the probability of all infinite-length strings is given by $\prod_{t=1}^{\infty} \left(1 - p_t\right)$.
However, by a standard result \citep[see e.g.,][Ch. 12]{Knopp1990TheoryAA}, we have that $\prod_{t=1}^\infty \left(1 - p_t\right) = 0 \iff \sum_{t=1}^\infty p_t \rightarrow \infty$, provided $p_t < 1$.\looseness=-1
\end{itemize}
Both cases together complete the proof.
\end{proof}

We can now see
that language processes are strictly more general than language models: \cref{eq:local-norm} shows us that any language model can be written as a language process, but \cref{prop:conversion} shows the converse is not necessarily true.
Indeed, \cref{prop:conversion} allows us to easily construct a simple language process (example given below) that cannot be converted to a language model, which motivates the formalism.
\begin{myexample}
Let $\vocab = \{a\}$.
Define a language process
$\boldY = \{Y_t\}_{t=1}^\infty$ over $\vocab$ such that each $Y_t$ is distributed according to $p(a \mid \yy_{<t}) = 1 - \frac{1}{2^{t+1}}$ and $p(\eos \mid \yy_{<t}) = \frac{1}{2^{t+1}}$. Note that we keep the convention that $Y_t=\bos$ for $t\leq 0$, and thus $p_0=0$. 
We have $\sum_{t=1}^\infty p_t = \frac{1}{2} < \infty$, so, by \Cref{prop:conversion}, $\boldY$ is not a language model.
Computing the infinite product $\prod_{t=1}^\infty \left(1 - p_t\right)$ shows $\boldY$ leaks $\approx .58$ to infinite strings.\looseness=-1
\end{myexample}

\paragraph{Life after $\eos$?}
\Cref{prop:conversion} further hints at the more intuitive difference between language models and language processes---what happens after $\eos$?
In the traditional definition of a language model (\Cref{defin:language-model}), life ends at $\eos$. 
That is, any string with symbols after $\eos$ would not be a valid sample from a language model because such strings are not in the model's support. 
On the other hand, a language process offers a more chipper view: once we hit $\eos$, we can just generate another symbol.
A language process is better thought of as an infinite babbler than a distribution over any sort of strings.
At some level, this is indeed the implicit view that is adopted by some when language modeling, as many language models do not have $\eos$ in the traditional sense. For the rest of this paper we will also take this view, and consider language processes for which we can continue generating after sampling an $\eos$ symbol.

\subsection{Other Useful Properties}
Next, we discuss some other properties about language processes that are important for understanding the theoretical results presented in \cref{sec:a-bit}.\looseness=-1
\begin{definition}[Markov]
A language process $\boldY = \{Y_t\}_{t=1}^\infty$ over alphabet $\vocab$ with distribution $p$ is Markov\footnote{Also known as a Markov chain.} if the following equality holds
\begin{align}
    p(Y_t \mid &\,\boldY_{<t}) = p(Y_t \mid Y_{t-k},\ldots,Y_{t-1}) \nonumber
\end{align}
where $k \geq 0$ is the Markov order. We again  take $Y_{t}$ for $t \leq 0$ to be $\bos$, indicating our initial distribution.\looseness=-1
\end{definition}
Many language processes are explicitly defined
to be Markov, e.g., ones based on $n$-gram language models.
However, many language processes based on recurrent neural networks are, in principle, non-Markov.
Yet despite being capable of learning non-Markov distributions, researchers have found that recurrent neural language models tend to learn Markov distributions. 
For instance, \citet{khandelwal-etal-2018-sharp} show that a recurrent neural language model's memory is empirically bounded at roughly 200 words.
Thus, we can still generally assume this property when working with language processes parameterized by such models.\footnote{Note that, in principle, human language is \emph{not} Markov, in so far as many linguists believe human language is capable of arbitrarily deep center-embeddings \cite{Chomsky57a,chomsky1995minimalist}. Yet research suggests that humans do not make use of this property in practice \cite{reich,Karlsson+2010+43+68}, and so we do not consider the Markovian property of most models as a limitation to their ability to model natural language in practice.\looseness=-1 }

\begin{definition}[Stationarity]
A $k$-Markov language process $\boldY = \{Y_t\}_{t=1}^\infty$ over alphabet $\vocab$ with distribution $p$ is \defn{stationary} if the following holds\looseness=-1
\begin{align}
  p(Y_{t+n} \mid &\,Y_{t-k+n},\ldots,Y_{t-1+n}) \\
  &= p(Y_t \mid Y_{t-k},\ldots,Y_{t-1}) \nonumber
\end{align}
for $n \geq 0$. We again  take $Y_{t}$ for $t \leq 0$ to be $\bos$, indicating our initial distribution.\looseness=-1
\end{definition}
While not theoretically Markovian, human language is generally considered stationary, i.e., the probability distribution over the next word should not depend on absolute position, but rather the history.\looseness=-1
\begin{definition}[Ergodicity]
A  language process $\boldY = \{Y_t\}_{t=1}^\infty$ is \defn{ergodic} if its statistical properties (e.g., ensemble averages) can be deduced from a single, sufficiently long, random sample.\looseness=-1
\end{definition}
The above definition is intentionally informal because ergodicity is a complex topic that would take time to treat thoroughly, but see \citet{mcmillan} and \citet{breiman} for a more rigorous discussion. 
One of the important implications of ergodicity for language processes, however, is rather straightforward: if our language process is over alphabet $\vocab$ with distribution $p$ and is ergodic, then for every symbol $y \in \vocab$ and for every history $\yy_{<t} \in \vocab^*$, there must exist an extension $\yy_{<t'} = \yy_{<t}, y_t ,\cdots, y_{t'\!-\!1}$ such that $p(Y_{t'} = y \mid \boldY_{<t'} = \yy_{<t'}) > 0$. 
In plain terms, this just says that we can always reach every word in our alphabet via some path no matter where we currently are. 
%
In our context, ergodicity also relates to the problem with $\eos$.
If we convert a language model into a language process (as discussed in \cref{sec:language_model}) and make the $\eos$ state absorbing,\footnote{This would be done by  setting the transition probability $p(Y_{t} = \eos \mid \boldY_{<t} = \yy_{<t}) = 1$ if $y_{t\!-\!1} = \eos$.} this language process must be non-ergodic, as once it encounters $\eos$, no other state is reachable.

\subsection{Estimating a Language Model from Data}\label{sec:estimation}
Language models are typically estimated
from language data.
The standard method for estimating the parameters of $p$ is via maximization of the log-likelihood of a training corpus $\mathcal{S}$\looseness=-1
\begin{align}\label{eq:nll}
        L(\vtheta;\mathcal{S}) = -\sum_{\yy \in \mathcal{S}} \sum_{t=1}^{|\yy|} \log p(y_t \mid \yy_{<t})
\end{align}
where $\vtheta$ are the model $p$'s parameters. The above is equivalent to minimizing the cross-entropy loss between $p$ and the empirical distribution. Note that we assume all $\yy \in \mathcal{S}$ end in the special $\eos$ token. 

\section{Information-Theoretic Properties of Language Processes}\label{sec:a-bit}
The view of language modeling as a discrete stochastic process naturally lends itself to an analysis through the lens of information theory.
Indeed, much of information theory is concerned with the study of discrete stochastic processes \cite[see e.g.,][Ch. 4]{cover2012elements}.
In this section, we review standard information-theoretic definitions in \Cref{sec:typicality} and build on these to introduce our own notion of local typicality in \Cref{sec:local-typicality}.

\subsection{Typicality}\label{sec:typicality}
An important definition in the study of stochastic processes is entropy rate, which generalizes the notion of entropy from a random variable to a stochastic process.\looseness=-1
\begin{definition}[Entropy Rate]\label{defin:entropy-rate}
Let $\boldY = \{Y_t\}_{t=1}^\infty$ be a stationary, ergodic discrete stochastic  process over alphabet $\vocab$ with distribution $p$.
The \defn{entropy rate} of $\boldY$ is defined as\looseness=-1
\begin{equation}
\ent(\boldY) \defeq \lim_{t\rightarrow \infty} \frac{1}{t}\ent(Y_1 ,\dots, Y_t)
\end{equation}
\end{definition}
The entropy rate is useful in that it tells us, in the limit, how spread out, i.e., entropic, the distribution is. Another interpretation is that it quantifies the complexity of $\boldY$. 
In the case of an i.i.d. process, the entropy rate and the entropy coincide, making the entropy rate a true generalization of the entropy.
Using entropy rate, we can define the notion of the typical set.

\begin{definition}[Typical Set]\label{defin:typical-set}
Let $\boldY = \{Y_t\}_{t=1}^\infty$ be a stationary, ergodic discrete stochastic process where each $Y_t$ follows distribution $p$ and takes on values in a finite support $\calY$. 
For $1\leq T<\infty$, the $(T,\varepsilon)$-\defn{typical set} of $\boldY$ is
the set of all sequences of length exactly $T$ with average per-symbol negative log-probability close to $ \ent(\boldY)$, i.e.
\begin{equation}
    \mathcal{T}_\varepsilon^{(T)} = \Big\{ \yy \mid \left|\frac{\log p(\yy)}{T} + \ent(\boldY)\right| < \varepsilon \Big\}
\end{equation}
\end{definition}
In informal terms, the typical set is the set of all samples that we would expect when sampling from $p$. To give the reader intuition about typicality, we now turn to a classical example.\footnote{See \citet{dieleman2020typicality} for further discussion of the concept of typicality in the context of generative modeling.}
\begin{myexample}
Consider an i.i.d. stochastic process $\boldY = \{Y_t\}_{t=1}^\infty$ where $Y_t$ is defined as the outcome of flipping a biased coin: we have $p(\textsc{heads}) = .6$ and $p(\textsc{tails}) = .4$. 
If we flip 100 coins, the most likely outcome is the sequence of 100 heads. 
However, this would be a surprising outcome to most people, who  would intuitively expect the sequence to consist of roughly $60\%$ heads and $40\%$ tails. Indeed, even for relatively large $\varepsilon$, the sequence of 100 heads is not in the  $\mathcal{T}_\varepsilon^{(T)}$ typical set; its average symbol probability is $.6 \gg 2^{-\ent(Y_t)} \approx 0.51$. 
\end{myexample}

The above example demonstrates that the typical set often does \emph{not} contain the most likely sequence. Additionally, the typical set is interesting because, as $T\rightarrow \infty$, it contains nearly all the probability mass; we formalize this property in a proposition.
\begin{proposition}\label{prop:typical-set-mass}
Let $\boldY = \{Y_t\}_{t=1}^\infty$ be a stationary, ergodic discrete stochastic process where each $Y_t$ follows distribution $p$ and takes on values in a finite support $\calY$. For every $\varepsilon > 0$, for sufficiently large $T$, the following conditions hold:
\begin{enumerate}[label={\roman*)}]
    \item $\sum_{\yy \in \mathcal{T}_\varepsilon^{(T)}}p(\yy) > 1 - \varepsilon$
    \item $(1-\varepsilon)2^{T(\ent(\boldY) - \varepsilon)} \leq |\mathcal{T}_\varepsilon^{(T)}| \leq 2^{T(\ent(\boldY) + \varepsilon)}$
\end{enumerate}
In words, as we take $T\rightarrow\infty$, the probability mass covered by the typical set is nearly 1 and the number of elements in it is nearly $2^{T\cdot \ent(\boldY)}$.
\end{proposition}
\begin{proof}
See \citet{breiman} for proof.
\end{proof}

\paragraph{What's wrong with the typical set?}
Let $\boldY$ be a stationary, ergodic language process. 
By the conditions of \Cref{defin:typical-set}, we know that $\boldY$ has a typical set.
We have motivated the typical set, intuitively, as the subset of strings that are usual or typical among all strings.
Under this intuition, it makes sense that---when using $\boldY$ as a language generator---this is the set from which we would like to select a string. 
A relatively straightforward corollary of \Cref{prop:typical-set-mass} is that ancestral sampling should pull from just this set. To see this, we can turn to  i) in \cref{prop:typical-set-mass}: since ancestral sampling provides an i.i.d. sample from $\boldY$, the probability of getting an element \emph{not} in $\mathcal{T}_\varepsilon^{(T)}$ as $T \rightarrow \infty$ is $(1-\varepsilon)$, i.e., practically never.
However, there is the confound that our models are not perfect representations of the true distribution behind the ``human'' natural language process.
Perhaps for this reason (and the reasons discussed in \cref{sec:communication}),
ancestral sampling is not known to result in samples that humans judge to be high quality in the task of language generation; rather it often leads to text that humans perceive as incoherent \cite{holtzman_curious_2020}. 
Furthermore, the typical set's definition relies on $\boldY$ being a stationary and ergodic language process.
As we saw in \cref{sec:process}, however, a language model that we convert into a language process will be non-ergodic by definition (at least if we keep $\eos$ as an absorbing state).
Thus, while the typical set is a natural starting point, it does not actually get us to our end goal of defining a set of strings that humans would find typical.
To remedy this problem, we introduce the new concept of local typicality.\looseness=-1

\subsection{Local Typicality}\label{sec:local-typicality}
A core contribution of this work is to define a more restrictive notion of typicality---termed here \defn{local typicality}---which we subsequently motivate as useful in the context of describing the set of strings humans typically produce.\looseness=-1

\begin{definition}[Locally Typical Set]\label{def:local-typical}
Let $\boldY = \{Y_t\}_{t=1}^\infty$ be a discrete stochastic process over finite support $\calY$.
The $(T,\varepsilon)$-\defn{locally typical set} of $\boldY$ is
the set of all sequences of length exactly $T$ such that\looseness=-1
\begin{align}\label{eq:local-set}
\mathcal{L}_\varepsilon^{(T)} &= \Big\{ \yy = y_0\cdots y_T \mid \forall  1 \leq t \leq T,\\
    & \Big|\log p(y_t \mid \yy_{<t}) + \ent(Y_t \mid \boldY_{<t} = \yy_{<t}) \Big| < \varepsilon \Big\} \nonumber
\end{align}
\end{definition}
In comparison to the typical set, the locally typical set further restricts the set of samples to those for which each individual symbol $y_t$ has probability near the local conditional entropy, i.e., the entropy of the distribution $p(\cdot \mid \yy_{<t})$. 
In general, there is no strong theoretical relationship between the typical set and the locally typical set.
However, in the case of an i.i.d. stochastic process we can prove that the latter constitutes a subset of the former.
\begin{proposition}\label{prop:iid}
Let $\boldY = \{Y_t\}_{t=1}^\infty$ be an i.i.d. discrete stochastic process, 
then $\mathcal{L}_\varepsilon^{(T)} \subseteq \mathcal{T}_\varepsilon^{(T)}$.\looseness=-1 
\end{proposition}
\begin{proof}
Since $\boldY$ is i.i.d., we have that $\ent(\boldY) = \ent(Y_t \mid \boldY_{<t}) = \ent(Y_t)$. 
Let $\yy$ be an element of $\mathcal{L}_\varepsilon^{(T)}$.
Then, $\sum_{t=1}^T \Big|\log p(y_t) + \ent(Y_t) \Big| <  T\varepsilon$.
Thus, by the triangle inequality,
$\Big|\sum_{t=1}^T \log p(y_t) + T \ent(Y_t) \Big| < T\varepsilon$, which implies $\Big|\frac{\sum_{t=1}^T \log p(y_t)}{T} + \ent(Y_t) \Big| < \varepsilon$, which implies $\yy \in \mathcal{T}_\varepsilon^{(T)}$.
\end{proof}

A natural question to ask at this point is why the definition of local typicality is useful in the context of a language process.
Our argument, presented in the following section, is cognitive in nature.

\section{Local Typicality in Natural Language}\label{sec:communication}

To motivate our definition of local typicality in the context of natural language, we must first look at language through an information-theoretic lens. 
We will consider \emph{two} distributions in this section: $\ptrue$, the distribution that a speaker of the language is assumed to generate strings from, and $\pmodel$ our language process that approximates $\ptrue$---albeit, perhaps not perfectly.
In this setting, we view a natural language string $\yy$ as a means of communicating some information, where each word $y_t$ is a symbol via which we construct our message. 
The information content of $\yy$ is then  defined as its negative log-probability under a specified distribution: $-\log \ptrue(\yy)$. 
Following the chain rule of probability, this quantity can be decomposed over words, i.e., the information content of a word is its negative log-probability given prior context: $-\log \ptrue(y_t\mid\yy_{<t})$.\looseness=-1

\subsection{Properties of Human Communication}
Given the above definitions, we can now ask a question at the heart of this work: what are the information-theoretic characteristics of natural language typically produced by humans.
In other words, what do strings sampled from $\ptrue$ look like, from the perspective of $\pmodel$, our trained language process?
Research in psycholinguistics suggests that a core component of what makes text human-like is its per-unit information content. 

To motivate this conclusion, we first consider a language user's objective. When using natural language, humans aim to transmit information efficiently while also minimizing the risk of miscommunication \citep{zipf1949human}. 
In order to achieve this goal, speakers avoid producing words with either very high or very low information content \citep[\emph{inter alia}]{fenk1980konstanz,aylett-turk,levy2007speakers,mahowald2013info},
a behavior inline with theories of efficient and robust communication.\footnote{See \citet{gibson2019efficiency} for an in-depth review of how efficiency has shaped the evolution of language.}
Indeed, cross-linguistic research has shown that languages trade-off information content and speech rate, perhaps aiming at a specific (optimal) information rate \citep{coupe2019different,pimentel+al.emnlp2021a}. 
Further, not using words in a context where they have very high or low information content avoids characteristics that appear to negatively impact traditional grammaticality judgments: an ideal natural language string  would not compensate for unusually near-zero probability in the first half, e.g., syntactic error, with unusually high probability in the second half, e.g., especially frequent words \cite{Schutze2016,lau2017}.\clara{elaborate on this more?}
\begin{figure}
    \centering
    \includegraphics[width=\linewidth]{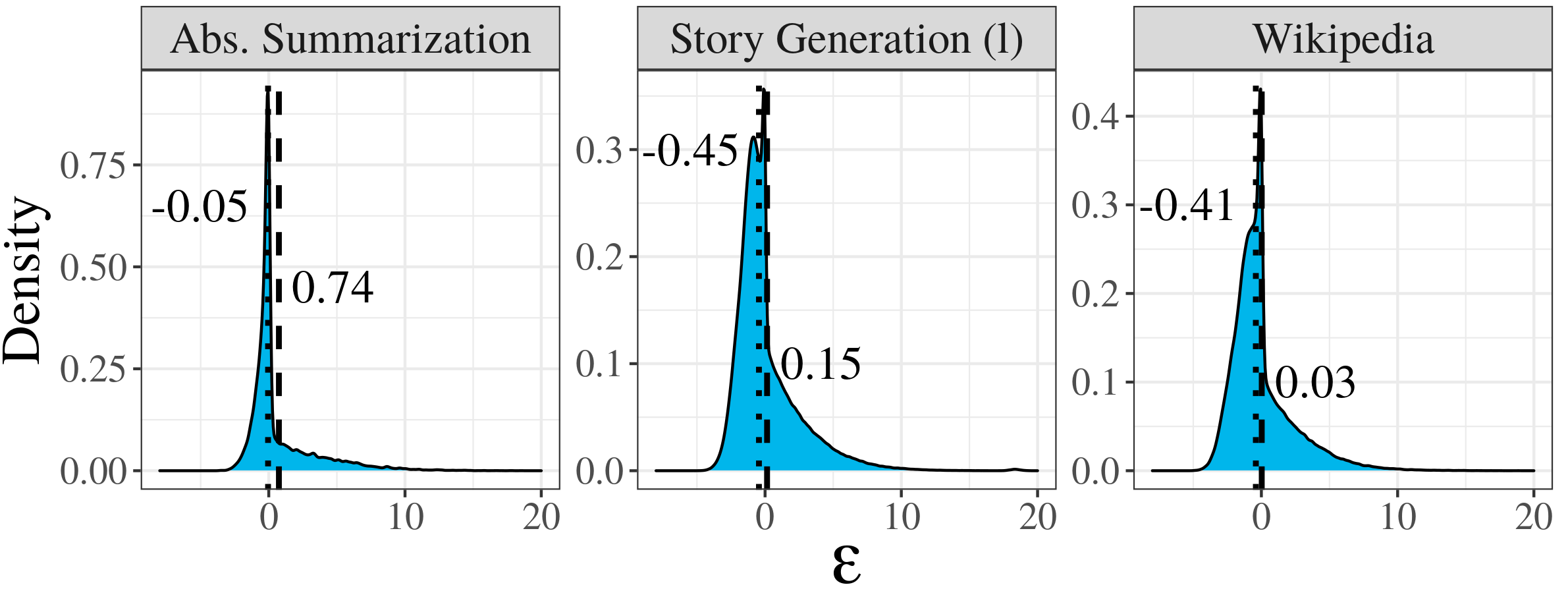}
    \caption{The per-token distribution of the deviation ($\varepsilon$) of information content from conditional entropy. Values are computed using the reference (human) text for three different language generation tasks, where probabilities and entropies are computed using probabilistic models trained on the respective task (see \cref{sec:exps} for model details). Dotted line and adjacent label indicate median $\varepsilon$ while dashed line and adjacent label indicate mean $\varepsilon$. 
    Per token distributions of conditional entropies and information contents  are shown in \cref{app:exp} for reference.
    \looseness=-1}
    \label{fig:reference_eps}
\end{figure}

\subsection{An Information-Theoretic Formalization}
The definition of local typicality presented in \cref{sec:local-typicality} can be viewed as an embodiment of the characteristics of human language just described above. 
One logical interpretation of these behaviors is that, at every time step, natural-sounding language should have per-symbol information content close to the expected (average) per-symbol information content.\footnote{The standard definition of (Shannon) entropy for a random variable $X$ with support $\mathcal{X}$ is equivalent to the expected information of $X$: $\ent(X) = -\sum_{x\in\mathcal X}p(x)\log p(x)$.}  
We formalize this relationship in the following hypothesis.\looseness=-1

\begin{hypothesis}\label{hyp:info-rate}
Samples $\yy = y_0 \cdots y_T $ from a human language process with distribution $\ptrue$ tend to belong to the process's locally typical set $\mathcal{L}_\varepsilon^{(T)}$ for large enough $T$ and some $\varepsilon > 0$.
In words, this means that we should expect every word in natural-sounding sentences to be close to the \emph{expected} information content under $\ptrue$, i.e., the conditional entropy given prior context.
\end{hypothesis}
We verify this relationship empirically using data from human language processes. 
In \cref{fig:reference_eps}, we show the distribution of the difference between the information content of $y_t$ and the \emph{expected} information content of $Y_t$, i.e., $-\log \pmodel(y_t \mid \yy_{<t}) - \ent(Y_t \mid \boldY_{<t} = \yy_{<t})$, according to the model on human-generated text.
The peaked nature of the distributions in  \cref{fig:reference_eps} reveals that human language indeed tends to have per-word information content quite close a specific value; the centering of these distributions around $\approx 0$ suggests that this value is $\ent(Y_t \mid \boldY_{<t} = \yy_{<t})$. 
Notably, \citet{meister+al.acl22a} shows the same is not true for text generated by models according to a number of different popular decoding schemes, which instead produce strings with much higher probability, i.e., with lower information content.\looseness=-1

In an ideal situation, such a property of natural language would be reflected in $\pmodel$, in which case sampling from the typical set should be sufficient to ensure human-like language.
However, our models are by no means perfect. 
The failure to capture the property of human language expounded in \cref{hyp:info-rate} may come from a number of possible modeling deficiencies, e.g., poor ability to capture the tails of these distributions. 
We hypothesize that, when using language models to generate text, enforcing this local-typicality criterion explicitly may serve as a patch for this shortcoming.

\section{Sampling from a Language Process}\label{sec:sampling}
In this section, we describe how to sample from a language process parameterized by the distribution $p$,\footnote{
Here we only consider locally normalized $p$, i.e., processes in which sampling is done on a word-by-word basis.} or in more commonly-used terminology, how to decode from $p$. 
There are many different algorithms one could employ to sample from $p$. 
The most intuitive strategy is ancestral sampling,\footnote{Another natural option would be to choose words which maximize the probability assigned by $p$ to the resulting string, but this work focuses on stochastic strategies.} which works as follows:  sample $y_t \sim p(\cdot \mid \yy_{<t})$ for each history $\yy_{<t}$ successively until some chosen criterion, e.g., the $\eos$ symbol is sampled or a maximum length is reached. 
Note that in the case of the former criterion, this procedure is equivalent to sampling entire strings according to the distribution $p$.
Perhaps the most popular set of techniques for sampling fall under  a paradigm we call \defn{truncated sampling}, where the vocabulary at a time step is truncated to a core subset of words.
For instance, \citet{fan_hierarchical_2018} propose limiting the sampling space to the top-$k$ most likely words in each decoding step, and \citet{holtzman_curious_2020} consider the smallest nucleus (i.e., subset) of words whose cumulative probability mass exceeds a chosen threshold $\eta$.\looseness=-1

In this paper, we give a general treatment of truncated sampling and then discuss our variant.
Given a context-dependent constraint subset $\mathcal{C}(\yy_{<t}) \subseteq \vocabeos$ of the vocabulary, we define the truncated distribution as\looseness=-1
\begin{align}
\pi(y \mid  \yy_{<t}) \defeq
\frac{p(y \mid  \yy_{<t}) \mathbbm{1}\{y \in \mathcal{C}(\yy_{<t})\}}{Z(\yy_{<t})}
\end{align}
where the normalizer is defined as 
\begin{equation}
    Z(\yy_{<t}) \defeq \sum_{y \in \mathcal{C}(\yy_{<t}) } p(y \mid \yy_{<t})
\end{equation}
and we call $\mathcal{C}(\yy_{<t})$ the truncation set. Now we give two examples of truncated samplers.
\begin{sampler}[Top-$k$ Sampling]
In  top-$k$ sampling, the truncation set $\mathcal{C}(\yy_{<t})$\ is defined as the top-$k$ highest-probability tokens $y$ according to $p(\cdot \mid \yy_{<t})$, i.e., the solution to the following subset maximization problem
\begin{equation}
\begin{aligned}\label{eq:topk-sampling-opt}
& \underset{\mathcal{C}(\yy_{<t}) \in \mathcal{P}(\vocabeos)}{\text{\normalfont maximize}}
& &  \sum_{y\in\mathcal{C}(\yy_{<t})}p(y \mid \yy_{<t})  \\
&\,\,\,\text{\normalfont subject to}
& & |\mathcal{C}(\yy_{<t})| \leq k
\end{aligned}
\end{equation}
where $\mathcal{P}$ is the power set operator.
\end{sampler}

\begin{sampler}[Nucleus Sampling]
In nucleus sampling, we choose a threshold parameter $\eta$ and define the truncation set $\mathcal{C}(\yy_{<t})$ as the solution to the following subset minimization problem
\begin{equation}
\begin{aligned}\label{eq:nuclear-sampling-opt}
& \underset{\mathcal{C}(\yy_{<t}) \in \mathcal{P}(\vocabeos)}{\text{\normalfont minimize}}
& &  |\mathcal{C}(\yy_{<t})|  \\
&\,\,\,\text{\normalfont subject to}
& & \sum_{y\in\mathcal{C}(\yy_{<t})}p(y \mid \yy_{<t}) \geq \eta 
\end{aligned}
\end{equation}
where again $\mathcal{P}$ is the power set operator.
\end{sampler}

\subsection{Shortcomings of Existing Algorithms}

To motivate sampling based on the locally typical set, we must first better understand the shortcomings of current decoding strategies.
While strings generated using stochastic strategies may have lower probability according to $\model$, they often outperform those decoded using maximization-based strategies in terms of qualitative metrics. 
A number of recent works have tried to offer explanations for this phenomenon. Some have attributed it to a diversity--quality trade-off \cite{zhang_trading_2020,BasuRKV2021}, while others blame shortcomings of model architectures or training strategies \cite{welleck2019neural,li-etal-2020-dont}.\looseness=-1

Our analysis from \cref{sec:communication} inspires an alternative explanation, motivated by information theory and psycholinguistics, for why models that perform so well (in terms of metrics such as perplexity) can still exhibit such undesirable behavior when used to generate text. 
First, the connection between probability and information content may explain why high-probability text is often dull or generic \cite{holtzman_curious_2020,eikema_is_2020}; its low information content likely makes for boring, i.e., uninformative, text.
This connection also offers a potential explanation for the rather strange behavior that, when a string has a repetitive loop, language models often assign increasingly higher probability to the repeated substring \cite{holtzman_curious_2020}; the substring conveys less and less information after each occurrence. 

A further implication of this framing is the equivalence between decoding strings from a probabilistic language generator and sampling messages from the natural language communication channel.  If we wish to solely sample from the subset of messages that a human would typically construct, i.e., that are human-like, then we should begin by narrowing down this subset to those messages that meet at least some of the same criteria as human-generated messages. In this work,  we have identified the criterion that such messages tend to be in the locally typical set. This observation motivates a new decoding strategy in which our information-theoretic criterion is explicitly enforced. 

\subsection{Locally Typical Sampling}

We now introduce our novel sampling algorithm, which we entitle \textbf{locally typical sampling}.
\begin{sampler}\label{alg:locally-typical-sampler}
Locally typical sampling is a truncated sampling scheme where the truncation set $\mathcal{C}(\yy_{<t})$ is the solution to the following subset optimization problem\footnotemark
\begin{align}
& \underset{\mathcal{C}(\yy_{<t}) \in \mathcal{P}(\vocabeos)}{\text{\normalfont minimize}}
 \sum_{y \in \mathcal{C}'(\yy_{<t})} \!\!\! \!\!\!\big|\ent(Y_t\mid \boldY_{<t} = \yy_{<t}) \\
 &\quad\quad\quad\quad\quad\quad\quad\quad\quad\quad\quad + \log p(y \mid \yy_{<t})\big| \nonumber  \\
&\,\,\,\text{\normalfont subject to}
 \sum_{y \in \mathcal{C}(\yy_{<t})}  p(y \!\mid\! \yy_{<t}) \geq \tau  \nonumber
\end{align}
\end{sampler}

In words, \Cref{alg:locally-typical-sampler} limits the sampling distribution to only those words with negative log-probability within a certain absolute range from the conditional entropy (expected information content) of the model at that time step.
In the spirit of nucleus sampling, this range is determined by a hyperparameter $\tau$, the amount of probability mass from the original distribution that we wish to consider.\looseness=-1

\footnotetext{\textit{Erratum}: This definition of the optimization problem being solved to produce the locally typical sampling truncation set allows for solutions in which continuations whose log-probabilities lie closest to the conditional entropy are excluded from the set. The authors are working on a new formulation that leads to solutions that better align with the greedy algorithm described subsequently in the computational complexity section---and therefore with the specification of the locally typical set in \cref{eq:local-set}. We thank Shay Cohen for pointing out the issue in our original formulation.} 

Interestingly, \Cref{alg:locally-typical-sampler} does \emph{not} imply that high-probability words should not be chosen.
Indeed, in the situation where conditional entropy is low, i.e., when the model places most of the probability mass on a small subset of words, it is likely the case that only high-probability words fall into the locally typical set.

\paragraph{Computational Complexity.}  From a practical perspective, locally typical sampling can be implemented with the same efficiency as nucleus or top-$k$ sampling. First, we compute the conditional entropy, which is an $\mathcal{O}(|\vocab|)$ operation. Second, we sort words by their absolute distance from $\ent(\model(\cdot \!\mid\!\boldY_{<t} =\yy_{<t}))$, which can be done in $\mathcal{O}(|\vocab|\log |\vocab|)$ time with standard sorting algorithms. 
Finally, we greedily take words from this list until their cumulative probability exceeds the threshold $\tau$, which again takes $\mathcal{O}(|\vocab|)$ time. Thus, creating our altered distribution has time complexity $\mathcal{O}(|\vocab|\log|\vocab|)$.\footnote{For each of the truncation sampling algorithms, the truncation set can also be identified using the selection algorithm (no sorting required) in $\mathcal{O}(|\vocab|)$ time. We provide the analysis using sorting as that is the standard implementation.\looseness=-1}\looseness=-1

\paragraph{Relationship to Other Decoding Strategies.}
Notably, we already see motivation for this criterion in the performance of several well-known decoding strategies. 
For example, beam search is the predominant decoding strategy for machine translation models \cite{Wu2016GooglesNM,edunov-etal-2018-understanding,ng_facebook_2019,meister_if_2020}, a setting in which  beam search (incidentally) often already enforces this criterion.\footnote{ When trained \emph{without} label-smoothing, which artificially inflates conditional entropies, machine translation models tend to have quite low conditional entropies \citep[see e.g., Fig. 3 in ][]{meister-etal-2020-generalized}. Therefore, at each decoding step, the set of words with negative log-probability near the conditional entropy of the model are typically those with high probability---the same as those chosen by beam search.} 
Yet, when used in more open-ended tasks, where the entropy of the language model is higher, beam search can lead to low-quality text \cite{li_diversity-promoting_2016,holtzman_curious_2020,welleck2019neural,meister+al.acl22a}. 
Locally typical sampling is also closely related to  nucleus sampling. 
When the probability distribution over the vocabulary has low conditional entropy, i.e., when there are only a few reasonable choices for the next word according to our model, nucleus and typical will have the same truncation set. Locally typical sampling and Mirostat \cite{BasuRKV2021} likewise have similar decision rules for truncation. Mirostat decodes strings such that they have a perplexity (or, equivalently, a per-word information content) close to a target value. 
In contrast to Mirostat, however, locally typical sampling does not require a specific target information content to be defined.
Rather, locally typical sampling defines this quantity as the conditional entropy, choosing it dynamically (per word) and making it less sensitive to hyperparameter choice. 
Finally, locally typical sampling is also related to \citeposs{pmlr-v119-braverman20a} strategy, which proposes a look-ahead decoding algorithm that generates text with a similar entropy rate to that of human-generated text.
Our strategy's motivation is similar---to match the tendencies in information content exhibited by human-generated text---albeit without requiring the computational overhead of a look-ahead strategy.

\begin{table*}[]
    \centering
    \small
    \begin{adjustbox}{max width=\linewidth}
    \begin{tabular}{llllllll}
        \toprule
        \multicolumn{8}{c}{\textbf{Story Generation}}\\
        & PPL (g) & PPL (i) & \textsc{Mauve} ($\uparrow$)& \rep ($\downarrow$)& Zipf  &$D$ ($\uparrow$)& Human  ($\uparrow$) \\
         \midrule
         Reference		& $	16.33	$&$	26.71	$&$	-	$&$	0.28	$&$	1.09	$&$	0.85	$&$	4.12\db{0.02}	$\\
\midrule																
Temperature	($\tau$=0.5)	& $	25.34\ddel{+9.01}	$&$	18.78\ddel{-7.93}	$&$	0.95	$&$	\mathbf{0.25}	$&$	\mathbf{1.07}\ddel{-0.02}	$&$\mathbf{0.87}	$&$	4.13\db{0.02}	$\\
Temperature	($\tau$=1)	& $	25.67\ddel{+9.34}	$&$	11.77\ddel{-14.94}	$&$	0.95	$&$	0.26	$&$	\mathbf{1.07}\ddel{-0.02}	$&$	\mathbf{0.87}	$&$	4.13\db{0.02}	$\\
Nucleus	($\eta$=0.9)	& $	7.75\ddel{-8.58}	$&$	10.25\ddel{-16.46}	$&$	0.95	$&$	0.35	$&$	1.29\ddel{+0.20}	$&$	0.79	$&$	4.09\db{0.02}	$\\
Nucleus	($\eta$=0.95)	& $	11.65\ddel{-4.68}	$&$	11.77\ddel{-14.94}	$&$	0.95	$&$	0.30	$&$	1.20\ddel{+0.11}	$&$	0.84	$&$	4.13\db{0.02}	$\\
Top-$k$	($k$=30)	& $	7.07\ddel{-9.26}	$&$	18.78\ddel{-7.93}	$&$	0.88	$&$	0.35	$&$	1.41\ddel{+0.32}	$&$	0.80	$&$	4.13\db{0.02}	$\\
Top-$k$	($k$=40)	& $	11.83\ddel{-4.5}	$&$	13.08\ddel{-13.63}	$&$	0.92	$&$	0.35	$&$	1.33\ddel{+0.24}	$&$	0.82	$&$	4.09\db{0.02}	$\\
Mirostat	($\tau$=3)	& $	8.14\ddel{-8.19}	$&$	\mathbf{23.53}\ddel{-3.18}	$&$	0.93	$&$	0.34	$&$	1.30\ddel{+0.21}	$&$	0.83	$&$	4.12\db{0.02}	$\\
Typical	($\tau$=0.2)	& $	\mathbf{14.25}\ddel{-2.08}	$&$	23.51\ddel{-3.20}	$&$	0.78	$&$	0.30	$&$	1.27\ddel{+0.18}	$&$	0.84	$&$	\mathbf{4.15}\db{0.02}	$\\
Typical	($\tau$=0.95)	& $	11.59\ddel{-4.74}	$&$	11.77\ddel{-14.94}	$&$	\mathbf{0.96}	$&$	0.31	$&$	1.21\ddel{+0.12}	$&$	0.84	$&$	4.13\db{0.02}	$\\
        \bottomrule
    \end{tabular}
    \end{adjustbox}
    \caption{Automatic quality and diversity metrics, as described in \cref{sec:setup},  along with human ratings on the  \textsc{WritingPrompts} dataset. Human ratings are averaged across criteria to form a single metric. Bolded values are the best results among decoding strategies, where for perplexity (PPL) and Zipf's coefficient, we take this to be the delta from measurements on human text (numbers in purple). Numbers in blue are standard error estimates. Results are from finetuned GPT-2 large.\looseness=-1
    }
    \label{tab:stories_large}
    \vspace{-7pt}
\end{table*}



\section{Experiments}\label{sec:exps}
In this section, we explore the efficacy of our decoding strategy on two natural language generation tasks: abstractive summarization and story generation. We assess performance with respect to several other stochastic decoding strategies: nucleus sampling, top-$k$ sampling, temperature sampling,\footnote{Temperature sampling is defined as ancestral sampling after local renormalization with an annealing term $\tau$.} beam search and Mirostat. Our evaluation includes both automatic metrics and human ratings.
 
\subsection{Setup}\label{sec:setup}
\paragraph{Models and Data.} We use the \href{https://huggingface.co/}{Hugging Face} framework \cite{wolf-etal-2020-transformers} for reproducibility, employing their implementations of nucleus, top-$k$, temperature sampling and beam search. We rely on the \href{https://github.com/basusourya/mirostat}{implementation} of Mirostat provided by its authors. For story generation, we finetune the  medium and large versions of GPT-2 \citep{radford_language_nodate} from checkpoints made available by OpenAI on the \textsc{WritingPrompts} dataset \cite{fan_hierarchical_2018}. We use the medium checkpoint finetuned on \textsc{WikiText-103} \cite{merity_pointer_2016} to produce the data used in \cref{fig:reference_eps}. For abstractive summarization, we use BART \cite{lewis_bart_2019} finetuned on the \textsc{CNN/DailyMail} dataset \cite{nallapati-etal-2016-abstractive}.\footnote{As we are interested in getting as close an estimate of $\emp$ as possible with our models $\model$, all fine-tuning is done \emph{without} label-smoothing. Note that label-smoothing may also artificially inflate conditional entropy estimates, as it pushes the learned distribution towards the most entropic distribution: the uniform distribution \cite{confidence_penalty}.}
All reported metrics are computed on the respective test sets.

\paragraph{Hyperparameters.} In a preliminary hyperparameter sweep using \textsc{Mauve}\footnote{We use the default settings given by the authors for all \textsc{Mauve} computations, albeit we employ different LMs in our parameter sweep vs. reported results (standard GPT-2 vs. GPT-2 large, respectively) to reduce bias in the final results. Notably, \textsc{Mauve} presents similar performances when used with these two pretrained LMs \citep{pimentel2022cluster}.} \cite{pillutla2021mauve}, we found $k=\{30,40$\}, $\eta=\{0.9,0.95\}$ and $\tau = 3.0$ to be the  best performing hyperparameters for top-$k$ sampling, nucleus sampling and Mirostat, respectively. For locally typical sampling, we found $\tau=0.2,\tau=0.95$ to provide the best results for story generation and abstractive summarization, respectively. Standard values according to the literature for other hyperparameters (i.e., for beam search and temperature sampling) were employed.
We use these values in our human evaluations and in computation of automatic metrics.\looseness=-1

\paragraph{Automatic Quality Metrics.}
As automatic quality metrics, we evaluate the generated text's perplexity---under both the model used to generate the text ($\mathrm{PPL}(g)$) and an independent, i.e., not finetuned, LM ($\mathrm{PPL}(i)$), namely GPT-2 large \cite{radford_language_nodate}. Several prior works have shown that neither low nor high perplexity \cite{zhang_trading_2020, nadeem_systematic_2020,pillutla2021mauve} are direct indicators of text quality. Rather, human-like text often has perplexity within a certain range. Consequently, we report the difference in this metric from the reference text as well. 
We additionally evaluate using \textsc{Mauve}\footnote{We use the \href{https://github.com/krishnap25/mauve-experiments.git}{implementation} provided by the authors.} \cite{pillutla2021mauve} with the reference text.

\paragraph{Automatic Diversity Metrics.}
We also evaluate locally typical sampling using automatic diversity metrics. We compute \rep \cite{welleck2019neural}, Zipf's coefficient, and $n$-gram diversity. For \rep we use the average of \rep/$\ell$ scores, as defined in Eq. 9 of \citep{welleck2019neural} for $\ell \in \{16, 32,128\}$. We define $n$-gram diversity $D$ as the average fraction of unique vs. total $n$-grams for $n \in \{1,2,3,4\}$ in a string\looseness=-1
\begin{equation}\label{eq:diversity}
D = \sum_{n=1}^4\frac{\# \textrm{unique $n$-grams in string}}{\# \textrm{ $n$-grams in string}}
\end{equation}
\begin{table*}[]
    \centering
    \small
    \begin{adjustbox}{max width=\linewidth}
    \begin{tabular}{llllllll}
        \toprule
        \multicolumn{8}{c}{\textbf{Abstractive Summarization}}\\
        & PPL (g) & PPL (i) & \textsc{Mauve} ($\uparrow$)& \rep ($\downarrow$)& Zipf  &$D$ ($\uparrow$)& Human  ($\uparrow$) \\
         \midrule
         Reference		& $	10.29	$&$	34.21	$&$	-	$&$	0.13	$&$	0.76	$&$	0.97	$&$	4.31\db{0.03}	$\\
\midrule																
Beam	($k$=5)	& $	1.39\ddel{-8.90}	$&$	\mathbf{34.21}\ddel{-0.00}	$&$	0.90	$&$	\mathbf{0.14}	$&$	\mathbf{0.77}\ddel{+0.01}	$&$	0.97	$&$	\mathbf{4.35}\db{0.03}	$\\
Temperature	($\tau$=0.5)	& $	\mathbf{7.10}\ddel{-3.19}	$&$	55.31\ddel{+21.1}	$&$	0.97	$&$	0.15	$&$\mathbf{	0.75}\ddel{-0.01}	$&$	0.97	$&$	4.25\db{0.03}	$\\
Temperature	($\tau$=1)	& $	6.46\ddel{-3.83}	$&$	35.96\ddel{+1.75}	$&$	0.95	$&$	\mathbf{0.14}	$&$	\mathbf{0.75}\ddel{-0.01}	$&$	0.97	$&$	4.29\db{0.03}	$\\
Nucleus	($\eta$=0.9)	& $	2.97\ddel{-7.32}	$&$	33.63\ddel{-0.58}	$&$	0.90	$&$	0.17	$&$	0.93\ddel{+0.17}	$&$	0.96	$&$	4.26\db{0.03}	$\\
Nucleus	($\eta$=0.95)	& $	3.96\ddel{-6.33}	$&$	56.43\ddel{+22.22}	$&$	\mathbf{0.99}	$&$	0.15	$&$	0.91\ddel{+0.15}	$&$	0.97	$&$	4.26\db{0.03}	$\\
Top-$k$	($k$=30)	& $	3.13\ddel{-7.16}	$&$	34.79\ddel{+0.58}	$&$	0.98	$&$	0.16	$&$	0.93\ddel{+0.17}	$&$	0.97	$&$	4.31\db{0.03}	$\\
Top-$k$	($k$=40)	& $	3.26\ddel{-7.03}	$&$	28.38\ddel{-5.83}	$&$	0.96	$&$	0.16	$&$	0.93\ddel{+0.17}	$&$	0.97	$&$	4.29\db{0.03}	$\\
Typical	($\tau$=0.2)	& $	3.80\ddel{-6.49}	$&$	62.33\ddel{+28.12}	$&$	0.72	$&$	\mathbf{0.14}	$&$	0.91\ddel{+0.15}	$&$	0.97	$&$	4.27\db{0.03}	$\\
Typical	($\tau$=0.95)	& $	3.86\ddel{-6.43}	$&$	56.67\ddel{+22.46}	$&$	0.96	$&$	0.15	$&$	0.92\ddel{+0.16}	$&$	0.97	$&$	4.32\db{0.03}	$\\
        \bottomrule
    \end{tabular}
    \end{adjustbox}
    \caption{Automatic quality and diversity metrics, as described in \cref{sec:setup},  along with human ratings on the  \textsc{CNN/DailyMail} dataset. Human ratings are averaged across criteria to form a single metric. Bolded values are the best results among decoding strategies, where for perplexity (PPL) and Zipf's coefficient, we take this to be the delta from measurements on human text (numbers in purple). Numbers in blue are standard error estimates.}
    \label{tab:summ}
\end{table*}

 \begin{figure*}[h]
    \centering
    \includegraphics[width=\textwidth]{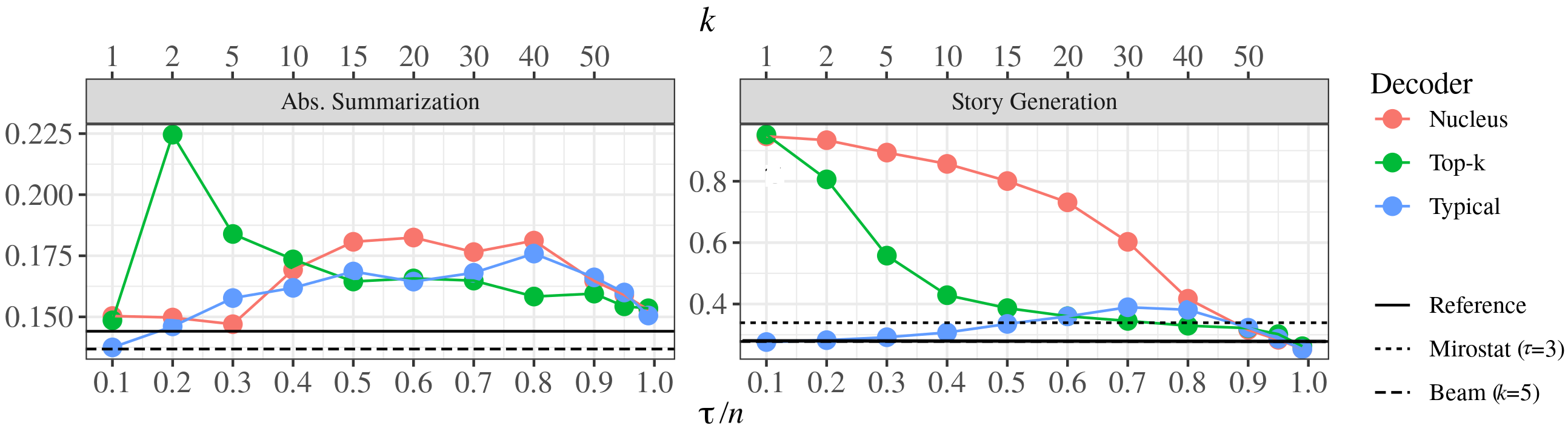}
    \caption{\rep \cite{welleck2019neural} values for different $k$ and $\tau$/$\eta$ (lower is better). Lines indicate \rep measurement for reference text and Mirostat (left)/beam search (right).}
    \label{fig:rep}
\end{figure*}
\begin{table*}[ht]
    \begin{center}
    \small
\begin{adjustbox}{max width=\linewidth}
    \begin{tabular}{p{0.1\linewidth}|p{0.8\linewidth}}
    \multicolumn{2}{c}{\textbf{\large Abstractive Summarization (CNN/DailyMail)}} \\
        \toprule
        
\textbf{\normalsize Prompt} &(CNN) The attorney for a suburban New York cardiologist charged in what authorities say was a failed scheme to have another physician hurt or killed is calling the allegations against his client ``completely unsubstantiated.'' Appearing Saturday morning on CNN's ``New Day,'' Randy Zelin defended his client, Dr. Anthony Moschetto, who faces criminal solicitation, conspiracy, burglary, arson, criminal prescription sale and weapons charges in connection to what prosecutors called a plot to take out a rival doctor on Long Island. ``None of anything in this case has any evidentiary value,'' Zelin told CNN's Christi Paul...\\
\midrule
\textbf{\normalsize Reference} & A lawyer for Dr. Anthony Moschetto says the charges against him are baseless.
Moschetto, 54, was arrested for selling drugs and weapons, prosecutors say.
Authorities allege Moschetto hired accomplices to burn down the practice of former associate.\\ [2pt]
\multirow{2}{*}{\shortstack{\bf \normalsize Beam \\ $k=5$}}& Dr. Anthony Moschetto faces criminal solicitation, conspiracy, burglary, arson and weapons charges. ``None of anything in this case has any evidentiary value,'' his attorney says.\\ [2pt]
\multirow{2}{*}{\shortstack{\bf \normalsize Nucleus \\ $\eta=0.95$}}& Dr. Anthony Moschetto, 54, pleaded not guilty to charges Wednesday. Two men -- identified as James Chmela and James Kalamaras -- were named as accomplices.\\ [2pt]
\multirow{2}{*}{\shortstack{\bf \normalsize Top-$k$ \\ $k=30$}}& Dr. Anthony Moschetto is accused of providing police with weapons and prescription drugs.
Authorities say he was part of a conspiracy to harm or kill a rival doctor.
His attorney calls the allegations against his client ``completely unsubstantiated'' \\ [2pt]
\multirow{2}{*}{\shortstack{\bf \normalsize Typical \\ $\tau=0.95$}} &Dr. Anthony Moschetto is charged with crimes including arson, conspiracy, burglary, prescription sale, weapons charges.
His attorney says ``none of anything in this case has any evidentiary value''\\
        \bottomrule
    \end{tabular}
     
     \end{adjustbox}
    \end{center}
   \caption{Sample generations for abstractive summarization; examples correspond to ID 1 in the test set. Decoding strategy hyperparameters are chosen based off of performance in human evaluations shown in \cref{tab:summ}. }
    \label{tab:abs_examples}
\end{table*}


\paragraph{Human Evaluations.}  
We use the \href{https://www.mturk.com/}{Amazon Mechanical Turk} framework to obtain human judgments of text quality from 5 different annotators on 200 examples per decoding strategy, per task. We use solely MTurk Master Workers in order to maximize the quality of our ratings. 
We follow \citet{delucia_decoding_2020} in setting up our evaluations. 
Each Human Intelligence Task (HIT) consists of either a single prompt from which a story should be generated or a single news article to be summarized. The raters are first presented with the different rating criteria, along with descriptions of the type of text that meets these criteria at different levels of the scale. Raters are additionally provided several examples of stories/summarizations that both meet and fail to meet the rating criteria. They are then presented with the respective prompt/news article and the corresponding stories/summaries generated by different decoders and by the reference in random order. For abstractive summarization, we ask annotators to score on \emph{fluency} and \emph{relevance}, while for story generation, annotators score on \emph{fluency}, \emph{coherence}, and \emph{interestingness}, each using a scale from 1 to 5. We choose these criteria following recommendations from \citet{van_der_lee_best_2019}.

For each story/summarization and each of the criteria, we take the median score across raters as the final score.\footnote{We use an attention check in each HIT. Responses where the attention check has been failed are thrown out.} 
Workers are paid \$1.50 per HIT for the abstractive summarization task and  \$2 per HIT for the story generation task, for which entries were longer. Note that these rates translate to $>$\$15/hour.\looseness=-1

\begin{table*}[ht]
    \begin{center}
    \small
\begin{adjustbox}{max width=\linewidth}
    \begin{tabular}{p{0.1\linewidth}|p{0.8\linewidth}}
    \multicolumn{2}{c}{{\large\textbf{Story Generation (WritingPrompts)}}} \\ 
\midrule
\textbf{\normalsize Prompt} &A kid doodling in a math class accidentally creates the world's first functional magic circle in centuries.\\
\midrule
\textbf{\normalsize Reference} & It was dark and Levi was pretty sure he was lying on his back. There was firelight flickering off of what was left of a ceiling. He could hear something but it was muffled. He was almost positive it was screaming. When he tried to move he felt an excruciating pain in his left side that caused him to cry out...\\ [2pt]

\multirow{2}{*}{\shortstack{\bf \normalsize Nucleus \\ $\eta=0.95$}}& ``Why do you draw a circle?'' He asked as he handed the piece of paper over to her.   She rolled her eyes and looked down at the picture, ``Because I like math!''   ``Then what?''   ``I like to draw the circles.'' She ran a hand through her curls and looked away. ``See, I have three circles.''   ``Hmph, I didn't think I saw that in your class.'' A younger student brought up the ceiling and murmured... \\ [2pt]

\multirow{2}{*}{\shortstack{\bf \normalsize Top-$k$ \\ $k=30$}}& ``Aha. Aah! That makes three.'' Professor Gosh sighed.   ``Oh my God, you had to have been there.''   ``Well, I'm not sure if that's the right term, but this circle is going to work.''   ``But I just told you that it wasn't working, didn't you, you little sh**? I'm not kidding! That circle is gon na work!''   ``So what now?'...\\ [2pt]

\multirow{2}{*}{\shortstack{\bf \normalsize Temp \\ $\tau=1.0$}} & It was teachers duty to ensure that our kingdom is pure and pure and successful, however it does not mean we do not try to be as rebellious as any other kingdom. Most of our teachers are suitable for rulership, being knowledgeable in sadistic rec thunking of time stopping magic circle created by several apprentices... \\[2pt]
\multirow{2}{*}{\shortstack{\bf \normalsize Mirostat \\ $\tau=3$}} & ready to go home, when a little kid starts to sketch a circle.   ``No. Just stop. It's not working.''   ``But the circle turns to smoke and then to a flame. It's not working, I know it is!'' Another kid, who was also drawing a circle with his phone, starts to scream. The teacher, looking at the circle, lost his patience and started to yell... \\ [2pt]
\multirow{2}{*}{\shortstack{\bf \normalsize Typical \\ $\tau=0.2$}} & As soon as he got the clock on his desk he jumped up. The noise it was making had changed his day forever. His hair, normally unkempt, now glistened and he had to have more hair because of it. ``Yes!'' He had declared triumphantly as he watched his father fumble around in the backpack in search of a little compass he always kept with him. He took it from the desk and quickly walked out the door to see what had happened. He wasn't the first child in the world to draw something...\\
        \bottomrule
    \end{tabular}
     
     \end{adjustbox}
    \end{center}
   \caption{Sample generations for story generation from GPT-2 large finetuned on the \textsc{WritingPrompts} dataset; examples correspond to ID 1 in the test set. Decoding strategy hyperparameters are chosen based off of best performance in human evaluations shown in \cref{tab:stories_large}. 
   Full stories shown in \cref{app:exp}.
   }
    \label{tab:summ_examples}
\end{table*}

\subsection{Results}

\paragraph{Quantitative Performance.} \Cref{tab:stories_large,tab:summ} show the results of our different evaluation metrics. Human scores are averaged across the qualitative metrics to give an aggregate score; the value in parentheses is the standard error of the estimate.  We show full breakdowns of score distributions in \cref{tab:breakdown}. We see that in general, locally typical sampling performs on par with or better than other sampling techniques, producing text with human quality ratings closest to that of the reference among the stochastic decoding strategies. Interestingly, beam search still outperforms locally typical sampling in abstractive summarization, albeit by a small margin. This could perhaps be attributed to the deterministic nature of beam search, which suggests that an interesting direction for future research may be a deterministic version of locally typical sampling, e.g., where the highest-probability word within the truncated set is always chosen. Importantly, all the strategies we explore are quite close to human-level performance---in some cases even surpassing human references in terms of ratings. At this level, it is perhaps only reasonable to expect that the differentiation between the top strategies is small. Accordingly, we also consider how robust locally typical sampling is to hyperparameter choice.
\cref{fig:rep} shows \rep measurements for different values of the hyperparameters $k$, $\eta$, and $\tau$ for top-$k$, nucleus, and locally typical sampling, respectively. Interestingly, \rep appears to be far less sensitive to $\tau$ than to $k$ and $\eta$. While many values of $k$ and $\eta$ appear to lead to degenerate repetitions in story generation, most values of $\tau$ lead to text with a \rep value on par with human text, demonstrating that an advantage of our technique is its robustness to hyperparameter choice. See \cref{fig:classification} in the appendix for a larger exploration of how other quality metrics vary as a function of $\tau$.\looseness=-1

\paragraph{Qualitative Performance.}
We present some examples of text generated according to each of the decoding strategies in \cref{tab:abs_examples,tab:summ_examples}. 
For both of the tasks, we choose the example with ID 1 in the respective test set and provide examples from each of the decoding strategies, employing the hyperparameter values that lead to the best human scores in \cref{tab:summ,tab:stories_large}. 
For the summarization task, we see that locally typical sampling provides a comprehensive and coherent summary of the article, quite similar to that of beam search. In comparison, the text produced by temperature sampling is not necessarily coherent; text from nucleus sampling and top-$k$ sampling misses some of the important information in the article,  e.g., the charges of burglary and arson.
While the qualitative performance in story generation is much more subjective, locally typical sampling arguably provides the most fluent story amongst all the decoding strategies. 
Other stories lack coherence and, even within the first few sentences, we see repeated phrases and words. 
Together, these results suggest that locally typical sampling may indeed produce more desirable text.

\section{Conclusion}
In this work, we analyze decoding from probabilistic language generators in the information-theoretic framework. 
We equate a language model to a discrete stochastic process\ryan{This isn't what we do. We formulate language modeling as one. The sampling itself is not a stochastic process.\response{clara}{I don't understand this} \response{ryan}{I would just rephrase it}\response{clara}{sure, but I wanted to understand why sampling isn't a stochastic process. Isn't that kind of the definition?}}, and use the known properties of such processes to quantitatively describe the samples we should expect.  
Motivated by results in psycholinguistics, we hypothesize that---with the goal of communicating efficiently and robustly---humans produce text whose per-word information content is within a close range of the \emph{expected} information content of a word given prior context. 
Current language models may fall short in capturing this property, which is a possible explanation for why the corresponding language processes often do not lead to human-like text. 
Yet, this observation provides a simple new criterion for decoding more human-like text from probabilistic language generators: constraining the sampling space to words that meet this criterion. In experiments on two language generation tasks, we find that our strategy---called locally typical sampling---leads to text of comparable or better quality than other stochastic decoding strategies according to human ratings. Further, when compared to these other decoding strategies, several quantitative properties of typically-sampled text more closely align with those of human text.

\section*{Acknowledgments}
We would like to thank Jason Eisner, Tim Vieira, Jennifer White and Ari Holtzmann for early conversations about the relationship between information theory and sampling. 
We would also like to thank Ehud Reiter, who served as our TACL action editor, and the the anonymous reviewers for their insightful feedback during the review process. Further, we are grateful to Eleanor Chodroff, Clément Guerner and Lucas Torroba Hennigen for their feedback on the manuscript of this work.

\section*{Ethical Concerns}
In order to complete our human evaluation, we used a crowdsourcing platform. 
For each task, we made sure that the crowdworkers would be paid (at minimum) a wage of \$15 per hour. 

Another ethical consideration worth discussing concerns the use of language models for text generation. 
Text generated by these models may contain malicious content, 
either by design of the user or as a byproduct of the training data/algorithm. 
While we hope the results of our work will not be misused, they may nonetheless provide insights for those employing these models with ill-intent as to how machine-generated text can be made more ``human-like,'' and thus more convincing.

\newpage
\clearpage
\bibliography{paper}
\bibliographystyle{acl_natbib}

\newpage
\appendix
\onecolumn

\section{Human Evaluation Setup}\label{app:setup}
\begin{figure}[H]
    \centering
    \includegraphics[width=\textwidth]{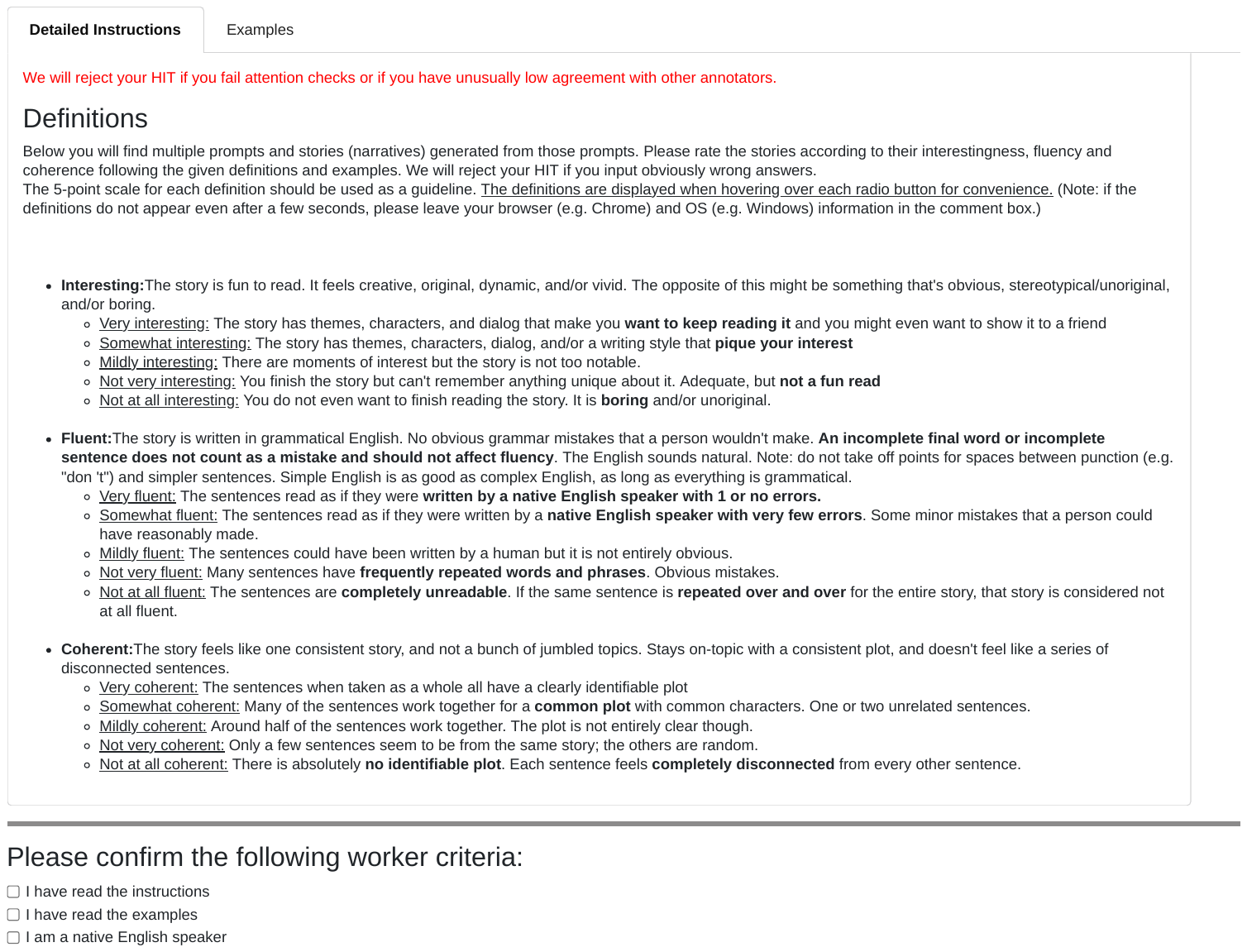}
    \caption{Stories survey.}\label{fig:stories}
\end{figure}

\begin{figure}[H]
    \centering
    \includegraphics[width=\textwidth]{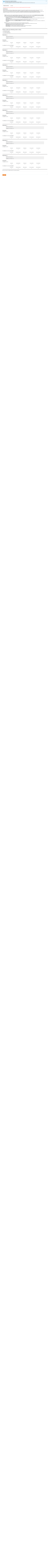}
    \caption{Summarization survey.}\label{fig:summarization}
\end{figure}

We use \href{https://www.mturk.com/}{Amazon Mechanical Turk} framework for collecting human ratings of text. We use solely MTurk Master Workers in order to maximize the quality of our ratings. 
For story generation and abstractive summarization, each Human Intelligence Task (HIT) consists of either a single prompt from which a story should be generated or a single news article to be summarized. The raters are first presented with the different rating criteria, along with descriptions of the type of text that meets these criteria at different levels of the scale. These definitions can be seen in \cref{fig:stories,fig:summarization}. Raters are additionally provided several examples of stories/summarizations meeting/failing to meet the rating criteria. They are then presented with the respective prompt/news article and the corresponding stories/summaries generated by different decoders and by the reference in random order. We use an attention check in each HIT. Responses where the attention check has been failed are thrown out. 
For each of the rating criteria, the rater assigns a score from 1 to 5. For each story/summarization and each of the criteria, we take the median score across raters as the final respective score. Statistics for these scores can be seen in \cref{tab:breakdown}. Workers are awarded \$1.50 per HIT for the abstractive summarization task and  \$2 per HIT for the story generation task, for which entries were longer. These rates translate to $>$\$15/hour. 
\section{Additional Results}\label{app:exp}

\begin{figure}[H]
    \centering
    \includegraphics[width=0.7\linewidth]{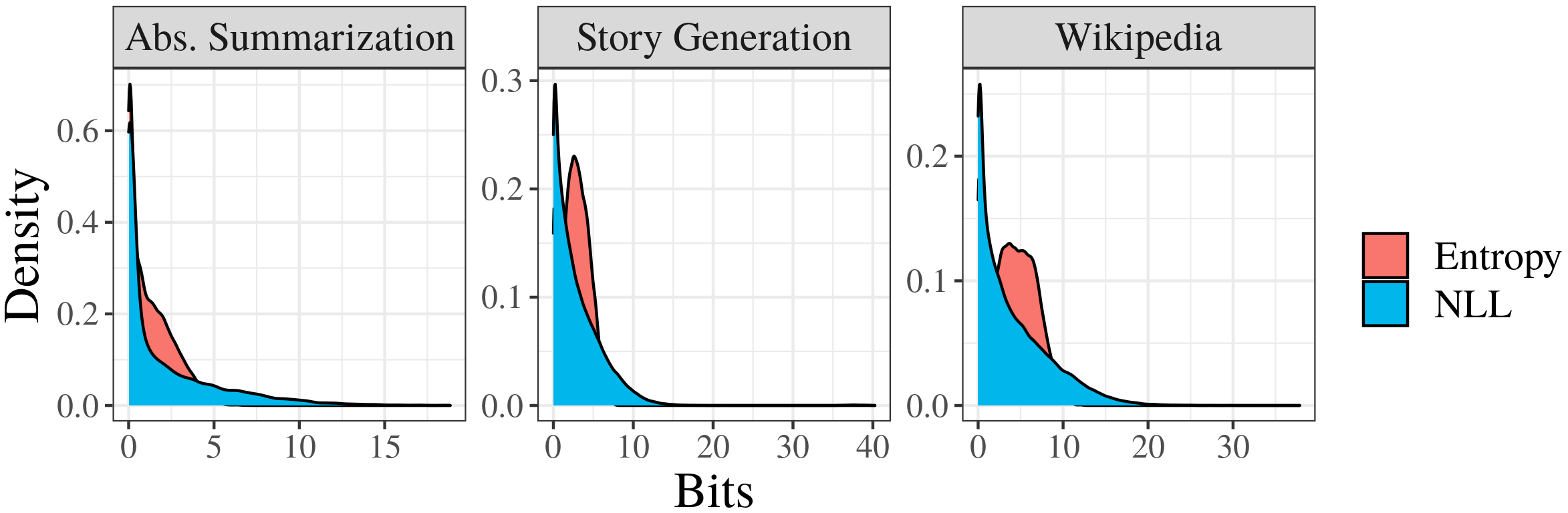}
    \caption{Distributions of conditional entropies and information contents per token for three different language generation tasks for human text, i.e., the reference text for each of the respective datasets.}
    \label{fig:reference_dists}
\end{figure}

\begin{table*}[h]
  \centering
  \small
  \begin{adjustbox}{max width=\textwidth}
  \begin{tabular}{ lrrr | rrr| rr }
   \toprule
  \multirow{2}{*}{{\bf Decoder}} &  \multicolumn{3}{c}{Story Generation (l)} &  \multicolumn{3}{c}{Story Generation (m)} & \multicolumn{2}{c}{Summarization} \\
  & Coherence & Fluency & Interestingness & Coherence & Fluency & Interestingness & Fluency & Relevance \\
  
  \hline\\[-1.5ex]

  Reference		& $	4.36\db{0.31}	$&$	4.25\db{0.23}	$&$	4.56\db{0.25}	$&$	4.02\db{0.27}	$&$	4.2\db{0.27}	$&$	4.15\db{0.2}	$&$	4.43\db{0.25}	$&$	4.18\db{0.27}	$\\
\midrule																		
Beam	($k$=5)	& $	-	$&$	-	$&$	-	$&$	-	$&$	-	$&$	-	$&$	4.47\db{0.24}	$&$	4.23\db{0.28}	$\\
Temperature	($\tau$=0.9)	& $	4.32\db{0.25}	$&$	4.16\db{0.19}	$&$	4.47\db{0.27}	$&$	4.02\db{0.22}	$&$	4.26\db{0.29}	$&$	4.19\db{0.24}	$&$	4.36\db{0.25}	$&$	4.13\db{0.26}	$\\
Temperature	($\tau$=1)	& $	4.36\db{0.28}	$&$	4.25\db{0.22}	$&$	4.47\db{0.30}	$&$	4.02\db{0.32}	$&$	4.2\db{0.29}	$&$	4.18\db{0.22}	$&$	4.42\db{0.26}	$&$	4.15\db{0.28}	$\\
Nucleus	($\eta$=0.9)	& $	4.32\db{0.25}	$&$	4.28\db{0.24}	$&$	4.48\db{0.31}	$&$	3.99\db{0.27}	$&$	4.16\db{0.32}	$&$	4.13\db{0.21}	$&$	4.39\db{0.27}	$&$	4.13\db{0.3}	$\\
Nucleus	($\eta$=0.95)	& $	4.3\db{0.28}	$&$	4.28\db{0.29}	$&$	4.49\db{0.26}	$&$	4.00\db{0.19}	$&$	4.24\db{0.35}	$&$	4.14\db{0.17}	$&$	4.44\db{0.26}	$&$	4.08\db{0.29}	$\\
Top-$k$	($k$=30)	& $	4.35\db{0.25}	$&$	4.21\db{0.24}	$&$	4.53\db{0.27}	$&$	4.03\db{0.24}	$&$	4.2\db{0.3}	$&$	4.16\db{0.22}	$&$	4.44\db{0.24}	$&$	4.18\db{0.26}	$\\
Top-$k$	($k$=40)	& $	4.34\db{0.27}	$&$	4.24\db{0.23}	$&$	4.53\db{0.25}	$&$	4.00\db{0.27}	$&$	4.17\db{0.31}	$&$	4.11\db{0.18}	$&$	4.41\db{0.25}	$&$	4.17\db{0.33}	$\\
Mirostat	($\tau$=3)	& $	4.39\db{0.27}	$&$	4.26\db{0.23}	$&$	4.55\db{0.27}	$&$	4.02\db{0.22}	$&$	4.16\db{0.32}	$&$	4.17\db{0.22}	$&$	-	$&$	-	$\\
Typical	($\tau$=0.2)	& $	4.36\db{0.29}	$&$	4.24\db{0.24}	$&$	4.55\db{0.25}	$&$	4.07\db{0.26}	$&$	4.23\db{0.32}	$&$	4.14\db{0.26}	$&$	4.37\db{0.28}	$&$	4.16\db{0.29}	$\\
Typical	($\tau$=0.95)	& $	4.35\db{0.28}	$&$	4.24\db{0.23}	$&$	4.53\db{0.26}	$&$	4.04\db{0.21}	$&$	4.18\db{0.31}	$&$	4.18\db{0.22}	$&$	4.42\db{0.28}	$&$	4.22\db{0.27}	$\\
    \bottomrule
  \end{tabular} 
  \end{adjustbox}
  \caption{Breakdown of human ratings on quality metrics per task; results for story generation are from finetuned versions of GPT-2 medium (m) and large (l). Values in blue are variances. }
  \label{tab:breakdown}
\end{table*}

\begin{figure*}[h]
\centering
    \includegraphics[width=\linewidth]{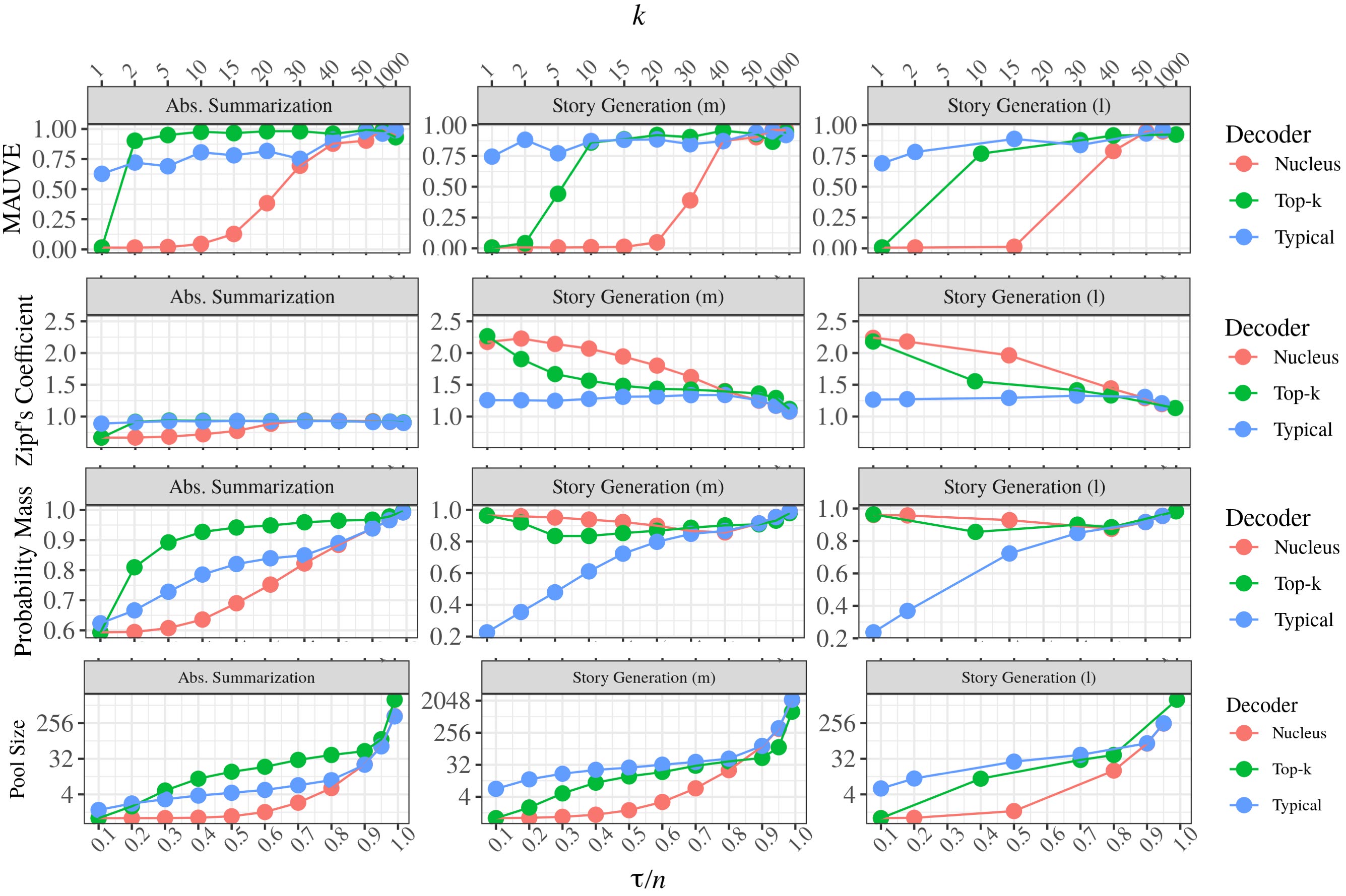} 
 
\caption{\textsc{Mauve}, Zipf's coefficient, (average) probability mass of candidate token pool and (average) candidate token pool size as a function of decoder hyperparameters for nucleus, top-$k$ and locally typical sampling.}
    \label{fig:classification}
    \vspace{-7pt}
\end{figure*}

\clearpage
\newpage
\begin{longtable}[H]{p{0.11\linewidth}|p{0.8\linewidth}}
    \multicolumn{2}{c}{\textbf{\large Abstractive Summarization (CNN/DailyMail)}} \\
        \toprule
\textbf{\normalsize Prompt} &(CNN) The attorney for a suburban New York cardiologist charged in what authorities say was a failed scheme to have another physician hurt or killed is calling the allegations against his client ``completely unsubstantiated.'' Appearing Saturday morning on CNN's ``New Day,'' Randy Zelin defended his client, Dr. Anthony Moschetto, who faces criminal solicitation, conspiracy, burglary, arson, criminal prescription sale and weapons charges in connection to what prosecutors called a plot to take out a rival doctor on Long Island. ``None of anything in this case has any evidentiary value,'' Zelin told CNN's Christi Paul.  "It doesn\'t matter what anyone says, he is presumed to be innocent." Moschetto,54, pleaded not guilty to all charges Wednesday.  He was released after posting \$2 million bond and surrendering his passport. Zelin said that his next move is to get Dr. Moshetto back to work. ``He's got patients to see. This man, while he was in a detention cell, the only thing that he cared about were his patients. And amazingly, his patients were flooding the office with calls, making sure that he was OK,'' Zelin said. Two other men -- identified as James Chmela, 43, and James Kalamaras, 41 -- were named as accomplices, according to prosecutors. They pleaded not guilty in Nassau County District Court, according to authorities. Both were released on bail. A requests for comment from an attorney representing Chmela was not returned. It's unclear whether Kalamaras has retained an attorney. Police officers allegedly discovered approximately 100 weapons at Moschetto's home, including hand grenades, high-capacity magazines and knives. Many of the weapons were found in a hidden room behind a switch-activated bookshelf, according to prosecutors. The investigation began back in December, when undercover officers began buying heroin and oxycodone pills from Moschetto in what was initially a routine investigation into the sale of prescription drugs, officials said. During the course of the undercover operation, however, Moschetto also sold the officers two semiautomatic assault weapons as well as ammunition, prosecutors said. Moschetto allegedly told officers during one buy that he needed dynamite to ``blow up a building.'' He later said he no longer needed the dynamite because a friend was setting fire to the building instead. Kalamaras and Chmela are believed to have taken part in the arson, according to prosecutors. ``The fire damaged but did not destroy the office of another cardiologist whose relationship with Dr. Moschetto had soured due to a professional dispute,'' according to the statement from the district attorney's office. Moschetto allegedly gave an informant and undercover detective blank prescriptions and cash for the assault and killing of the fellow cardiologist, according to prosecutors. He also requested that the rival's wife be assaulted if she happened to be present, authorities said. ``He was willing to pay \$5,000 to have him beaten and put in a hospital for a few months, and then he said he would pay \$20,000 to have him killed,'' said Assistant District Attorney Anne Donnelly, according to CNN affiliate WCBS.\\
\midrule
\textbf{\normalsize Reference} & A lawyer for Dr. Anthony Moschetto says the charges against him are baseless.
Moschetto, 54, was arrested for selling drugs and weapons, prosecutors say.
Authorities allege Moschetto hired accomplices to burn down the practice of former associate.\\ [5pt]
\textbf{\normalsize Beam }& Dr. Anthony Moschetto faces criminal solicitation, conspiracy, burglary, arson and weapons charges. ``None of anything in this case has any evidentiary value,'' his attorney says.\\ [5pt]
\textbf{\normalsize Temp}& Dr. Anthony Moschetto accused of planning to kill doctor, drug dealer. Scheme imploded when a crisis developed in their relationships.
Preliminary bond hearing set for August 5.\\ [5pt]
\textbf{\normalsize Nucleus }& Dr. Anthony Moschetto, 54, pleaded not guilty to charges Wednesday. Two men -- identified as James Chmela and James Kalamaras -- were named as accomplices.\\ [5pt]
\textbf{\normalsize Top-$k$}& Dr. Anthony Moschetto is accused of providing police with weapons and prescription drugs.
Authorities say he was part of a conspiracy to harm or kill a rival doctor.
His attorney calls the allegations against his client ``completely unsubstantiated'' \\ [5pt]
\textbf{\normalsize Typical} &Dr. Anthony Moschetto is charged with crimes including arson, conspiracy, burglary, prescription sale, weapons charges.
His attorney says ``none of anything in this case has any evidentiary value''\\
\midrule
    \multicolumn{2}{c}{{\large\textbf{Story Generation (WritingPrompts)}}} \\ 
\midrule
\textbf{\normalsize Prompt} &A kid doodling in a math class accidentally creates the world's first functional magic circle in centuries.\\
\midrule
\textbf{\normalsize Reference} & It was dark and Levi was pretty sure he was lying on his back. There was firelight flickering off of what was left of a ceiling. He could hear something but it was muffled. He was almost positive it was screaming. When he tried to move he felt an excruciating pain in his left side that caused him to cry out. His hand moved to it instinctively and found something protruding from the wound. It seemed to be a pipe of some kind. He attempted to pull it out but it was covered in blood making it slick.   Levi's ears began ringing and the sounds began to become clearer, it felt quite a bit like some one was driving needles into his eardrums. The sounds he was hearing were definitely screams and not just one person's, a lot of people were screaming or yelling. There was some one close to him that was crying. He looked in the direction of the tears and seen his sister kneeling beside him, her hands covering her face.   ``What happened Laur?'' He was shocked at the sound that barely rasped out from between his lips. His sister's hands jerked down and she stared down at Levi with a shocked look on her face. ``bu... I tho... you weren't breathing!'' What started as a whisper ended in yell as she threw her self across her brother and began to sob anew. Levi cried out hoarsely but she didn't hear. She just continued to cling to him like a little girl that had just found her lost doll.   He put one of his arms around her and scanned the room as much as he could. It looked like a warzone, like something out of one of the many shooters in his gaming collection. ``What the hell?'' There were dead bodies everywhere, he recognized some of them. There were firefighters and EMT's running around trying to find survivors in the rubble. Most of the screams were coming from survivors.   He seemed to be laying on top of the bottom of a desk, and he was pretty sure the pipe sticking out of his side was a actually one of the legs. Then he spotted it lying about two feet from his right leg, a round section of desk about the size of a softball. On it was a round symbol with geometric shapes glowing with dark red embers and a dull tendril of smoke rising up from it. It all came back to him in rush.   He drew that in his notebook. It was second period and his trig teacher had this monotonous voice that could put a crack head to sleep. Laurana caught him doodling and had thrown a pencil at him to scold him silently, which made him jerk as he was connecting the last line on his drawing. Then there was the light and the heat and lastly the dark. Did he do this? What the hell was going on here?   A voice brought him out of his little flashback. ``Sir. Sir? Sir!'' it was a firefighter. ``I need you to stay still. You're going to hear a loud noise from behind you. I need you stay calm.''  Kay. ``He just blew up his classroom with a piece of paper and a pencil and this guy wants him to stay calm.'' it's going to cause this desk to vibrate a lot and your going to feel some pain but there is nothing that we can do about it. We will work as fast as possible.''   The sound of the saw kicking on was not near as bad as he had thought, but Levi was no where near ready for when it made contact with the leg of the desk. His cries of agony began to drown out the sound of everything around him until darkness overwhelmed him.   Sorry about my grammar it's been a while since I have written anything also, first time poster so take it easy on me; p\\ [5pt]

\textbf{\normalsize Nucleus }& ``Why do you draw a circle?'' He asked as he handed the piece of paper over to her.   She rolled her eyes and looked down at the picture, ``Because I like math!''   ``Then what?''   ``I like to draw the circles.'' She ran a hand through her curls and looked away. ``See, I have three circles.''   ``Hmph, I didn't think I saw that in your class.'' A younger student brought up the ceiling and murmured.   ``Yeah, sure. It's just, why do we draw the circles?''   ``Because they're circles.''   She looked up. ``But they don't match up!'' She pointed at the nearby wall, ``They don't start out like this!'' She ran a hand over the map on the wall, ``It starts out the same.'' She turned and pointed to the map marked with candles. ``You have fire here.''   ``Well that's nice.'' He turned away from her, ``I don't think the artists in class thought about it that way.''   She pointed again.   ``See? Even your circle!''   ``That's my circle.''   ``Look, Mason.''   ``Hear, daughter.''   ``I don't see a man, and so can't prove it.''   The two young adults nodded in agreement. Then she began to speak, ``I draw the circles because I like to guess about what will happen when I make circles.''   ``Yeah, and that's what we're going to do when you make a circle and turn the paper to our faces.'' \\ [5pt]
\textbf{\normalsize Top-$k$}& ``Aha. Aah! That makes three.'' Professor Gosh sighed.   ``Oh my God, you had to have been there.''   ``Well, I'm not sure if that's the right term, but this circle is going to work.''   ``But I just told you that it wasn't working, didn't you, you little sh**? I'm not kidding! That circle is gon na work!''   ``So what now?''   ``Well, that's for you to figure out. First, you've got to figure out a way to communicate with the circle. Maybe the spell isn't right for it, but it's not going to ruin the circle.''   ``Okay, okay, you've told me to figure it out so I'll do a whole lot of work. But how exactly do you communicate a spell to a circle of stones in a basement?''   ``Simple, really. You put it inside the circle, and it magically works its way around the circle to get to the person you wrote'I AM' on.''   ``But how? Why am I supposed to be the main caster?''   ``Because you don't have magic powers to cast spells.''   ``What? That's not even supposed to be possible.''   ``You are special, you know.'' Professor Gosh looked at me with a little bit of sympathy.   ``I can't believe you're a wizard, you can do all this stuff when you're a kid.''   ``Well, yeah, that's what they told me, so I guess you're right. But no matter how hard you try, you can't control magic. Don't worry, we'll figure out a way to make you a wizard one of your own.''   ``Okay, I think you understand, I still haven't figured out how to create a spell, but I am so excited about the possibilities!''   ``Alright, let's do this.''\\ [5pt]

\textbf{Temp} & It was teachers duty to ensure that our kingdom is pure and pure and successful, however it does not mean we do not try to be as rebellious as any other kingdom. Most of our teachers are suitable for rulership, being knowledgeable in sadistic rec thunking of time stopping magic circle created by several apprentices at the same time.   After the first amendment, we decided that whilst having a society that thrives on protecting and serving the majority of people, whilst ensuring that the needs of some are met, the elements of freedom, equality and liberty shall not be infringed upon. Sadly, we ’ ve done little to uphold those values since the day our forefathers signed the ancient Papercut act and limited the extent of human freedoms. It is now a day we lay a steady hand on the pages of history, hoping that the rules shall be rewritten and brought to what they once were.   The next barrier to seeing the world anew as a black pantheon was a fallen occult student, Mr Pegarchow in Commerce, brought to history via the magic eye how to make a magic circle in the first place. This student made a hoop in the library that can be opened with a simple spiral. If a magic circle is inside this hoop, the magical illusion created is that a circle is made with the appropriate essence of spirit. An image of Pgyarchow being covered by the hoop, with the colours that inhabit it are verily magical. The magician essentially created… magic. To this day, not a single piece of paper hath survived contact with the magic eye.   There were so many experiments looking for what function to draw in, or when it starts, 3rd century of my not-so-distant time. Nothing turned out as expected and everyone quickly passed into subculture that embraced the forbidden art of forgetfulness. However, even through the strongest taboo that could cause, symbols of independent spirit were still made and used as religion source material.   I am the last wizard right now though not by choice, which was rather problematic after my mentor ’ s death. However, I am different, and a magician, not by birth but via training and knowledge dedcialized from the single book of knowledge stored in my heart, able to share it with everyone, even those who share their mothers blood.   To mhm, Lizonun \\[5pt]
\textbf{\normalsize Mirostat} & ready to go home, when a little kid starts to sketch a circle.   ``No. Just stop. It's not working.''   ``But the circle turns to smoke and then to a flame. It's not working, I know it is!'' Another kid, who was also drawing a circle with his phone, starts to scream. The teacher, looking at the circle, lost his patience and started to yell at the student: ``This circle is not working, you need to come up with some more inventive magic demonstration!''   ``I like it. You can draw it with your phone, it's not hard.''   ``No, stop it. Just stop drawing it like that and move on with your day. You look troubled.''   ``No, I do. I can get some ideas for my magic circle. Also it looks cool and I can tell everyone that it's working.''   ``Ok, but you still have to come up with a name for the circle.''   ``I can come up with kenning, kenning-perforant, kenning-untergrund or kenning-unterhrund depending on the layout. Then my circle is named kenning-unterhrund, and then I post that on reddit, and I sell it on Etsy!''   ``Uhh... you spelt it wrong again using your phone, kenning-unterhrund for what's a circle in the middle of a room! You're not getting any creativity from that!''   ``Still, I'm not that bad. I could come up with a name too. We could see if our circle works on reddit!''   The teacher starts to object, but the kid just brushes him off: ``open the door and let me in, I'll tell everyone my name for my circle.''   The door is open and the teacher comes in. ``KENNING-UNTER DOES N'T WORK, IT'SBAGS!''   The kid answers: ``Open the door and let me in. If there's no door, I'll come back tomorrow and make a new one.'' He sounds confident in his plan.   ``Ok, kenning-untergrund, kenning-unterhrund or kenning-unteromactually totalitarian need you! Honoring your name will make your circle work better!''   The kid is ready to spell it then: ``The kenning-untergrund, kenning-unterunterhrund, kenning-unteronAAAKE!''   The teacher asks: ``What do you mean fails?''   ``Ok, and I forgot why I just copy all my circles to my phone:)''   ``You can't make circles! That would be impossible. And you forgot the reason for your circle to not work:)''   ``Yes, I did! I forgot the purpose of my circle and this happened to me too:)''   ``That might happen, but get out of here immediately, the spell is about to hit the ground!''   The kid grabs the teacher by the shirt collar: ``So what did you get out of it? I thought magic circles are making people happy?''   The teacher looks at him with fury in his eyes: ``You should go to the yelling and crying section of the school, before the spell hits the ground. You're going to repeat the'Wizard' spell again.''   ``Thanks, but no, I'll leave.''   The kid grabs the ground and he leaves the school. Then his circle is gone.   The teacher is still trying to figure out what should be done about the circle because he forgot to ask some of his other students. So he goes home to his basement, to his computer and he starts to draw a circle. And then he draws another one. And another one. Soon enough, there are some things that are drawn on his screen: a circle with smoke coming out of it, a circle with fog coming out of it, a circle with fire coming out of it, a circle with flavouring of words coming out of it, a circle with adding words into it and a circle with some mixture of words coming out of it. As the circle is wants to be drawn, he starts to move his finger Mask he was using to draw the circle.   When suddenly, his finger don't move, slowly but steadily, like a hand is starting to move against his will. The students are looking at the screen now. They start to scream and yell: ``Why is the hand still moving?!!''   ``Yes, \\ [5pt]
\textbf{\normalsize Typical} & As soon as he got the clock on his desk he jumped up. The noise it was making had changed his day forever. His hair, normally unkempt, now glistened and he had to have more hair because of it. "Yes!" He had declared triumphantly as he watched his father fumble around in the backpack in search of a little compass he always kept with him. He took it from the desk and quickly walked out the door to see what had happened. He wasn't the first child in the world to draw something to him and so far, so good. It wasn't as big as a tree, but it did sparkle a little bit and made a few notes in his calculus textbook. As he left the room, the others around him also drew and explained to him their ability. Most kids who knew their way around magic started to teach their spells in a lesson on learning. His teachers tried to tell him about his parents but he only heard his name as they asked if he knew about them. They explained how he would always go with them when he learned something and then asked how he had found the way. His mother said she always did the work when she taught and then how his father used the ingredients he had brought along to create his circle. It wasn't anything like what his mother would use to do her magic but he wasn't interested in her or his magic because it wasn't his favorite type. As he got older he got to watch as the world moved. His family went out on field trips to look at stars and wonder where the other world is. They asked where his mother and father went, and how long they spent on that. It wasn't anything like he expected but they never explained what the field trips were about. As he grew older he became a man with more responsibility and began to notice things like this happen to the rest of the world. As a man who didn't believe in magic he found his circle was missing from the side of the road he always stopped on to check his bearings. His neighbors thought it was weird that his neighbors' circles would get in the way. One day his friends stopped on the street he was walking down and told him to get to his car and take him to a store so they could find his circle. When they found his circle they gave it back and it had no clue where it came from. After some further digging he discovered the missing part of the circle. A pentagram that appeared out of nothingness on the road and for the longest time, no one had been able to find it. A couple weeks later a truck came out and dumped all of the construction workers that were using the road and blocked it. When his circle had disappeared the news talked about the weird pentagram on the road.   This went on for months until his mother noticed he wasn't coming home. When she came home, she looked in his room and he wasn't there. When he wasn't there he never made his presence known to her and never tried to teach his spells to her. They eventually went back to their apartment together. Her father didn't even acknowledge his daughter at first but his wife didn't think to call him or get his attention. The father, seeing the pattern and a good time with his daughter decided to stay. She told her parents to check their calendar. It had stopped being so late that her father thought to take his place to try to sleep off his spell exhaustion. The parents realized something was up and the wife suggested she get her boyfriend and see if she could have a quiet time.   It wasn't a normal time to do such things, but when he found her and saw his father had passed he told his girlfriend that they would meet for drinks to get over it. The wife got ready to go, and they drove home. She pulled over in front of their house. A truck pulled up in the driveway, stopped and looked out and began to slowly pull into the driveway. When she was looking in his eyes, she asked if he knew about her mother's strange powers. When he replied he asked how his father got the circle that they kept around for all of these years. \\
& The man pulled up in his truck and gave his girlfriend the back seat and started the truck. She took him home, they drank, and watched a movie about dragons and aliens and made up a bedtime story for the boy and girl. After dinner the boy slept through the night. He dreamt that the circle had gone missing, that it had disappeared in his yard, and that the car parked next to him was covered in dust and it smelled like his room had been raided by monsters.   After his sleep, the mother took him to his room to wake him up. When she entered he didn't react at all. ``Are you alright honey?'' The mother asked him. The boy responded in the most puzzled tone she had ever heard him.\\
        \bottomrule
        \caption{Full sample generations for abstractive summarization and story generation. We use samples from the model fine-tuned from GPT-2 large for story generation.}
    \label{tab:examples_full}
\end{longtable}

\end{document}